\pdfoutput=1

\documentclass[10pt,twocolumn,letterpaper]{article}

\usepackage{wacv}              

\usepackage{graphicx}
\usepackage{amsmath}
\usepackage{amssymb}
\usepackage{booktabs}

\usepackage[pagebackref,breaklinks,colorlinks]{hyperref}
\usepackage[accsupp]{axessibility}

\usepackage[capitalize]{cleveref}
\crefname{section}{Sec.}{Secs.}
\Crefname{section}{Section}{Sections}
\Crefname{table}{Table}{Tables}
\crefname{table}{Tab.}{Tabs.}

\def\wacvPaperID{650} 
\def\confName{WACV}
\def\confYear{2025}

\usepackage{xspace}
\usepackage{svg}
\usepackage[inline]{enumitem}
\usepackage{caption}
\usepackage{subcaption}
\usepackage{makecell}
\usepackage{tikz}
\usepackage{pgfplots}
\usetikzlibrary{shapes.geometric,shapes.multipart,arrows,positioning,calc,shapes.arrows,arrows.meta,pgfplots.statistics,backgrounds,fit}
\usepgfplotslibrary{groupplots}
\tikzset{fit margins/.style={/tikz/afit/.cd,#1,
    /tikz/.cd,
    inner xsep=\pgfkeysvalueof{/tikz/afit/left}+\pgfkeysvalueof{/tikz/afit/right},
    inner ysep=\pgfkeysvalueof{/tikz/afit/top}+\pgfkeysvalueof{/tikz/afit/bottom},
    xshift=-\pgfkeysvalueof{/tikz/afit/left}+\pgfkeysvalueof{/tikz/afit/right},
    yshift=-\pgfkeysvalueof{/tikz/afit/bottom}+\pgfkeysvalueof{/tikz/afit/top}},
    afit/.cd,left/.initial=2pt,right/.initial=2pt,bottom/.initial=2pt,top/.initial=2pt}
\pgfplotsset{compat=1.18}

\newcommand{\ours}[0]{ElasticViT\xspace}

\newcommand{\afterplot}{\vspace{-.1in}} 

\DeclareMathOperator{\pred}{pred}
\DeclareMathOperator{\E}{E}
\DeclareMathOperator{\miss}{miss}
\DeclareMathOperator{\scale}{scale}
\DeclareMathOperator{\pos}{pos}
\DeclareMathOperator{\PE}{PE}

\definecolor{crimson2143940}{RGB}{214,39,40}
\definecolor{darkgray176}{RGB}{176,176,176}
\definecolor{darkorange25512714}{RGB}{255,127,14}
\definecolor{forestgreen4416044}{RGB}{44,160,44}
\definecolor{gray}{RGB}{128,128,128}
\definecolor{mediumpurple148103189}{RGB}{148,103,189}
\definecolor{steelblue31119180}{RGB}{31,119,180}

\begin{document}

\title{Beyond Grids: Exploring Elastic Input Sampling for Vision Transformers}

\author{Adam Pardyl$^{1,2,3}$ \qquad Grzegorz Kurzejamski$^{1}$ \qquad Jan Olszewski$^{1,4}$ \\
Tomasz Trzciński$^{1,5,6}$ \qquad Bartosz Zieliński$^{1,2}$ \\
$^1$IDEAS NCBR \\
$^2$Jagiellonian University, Faculty of Mathematics and Computer Science \\
$^3$ Jagiellonian University, Doctoral School of Exact and Natural
Sciences \\
$^4$ University of Warsaw \qquad $^5$ Warsaw University of Technology \qquad $^6$ Tooploox 
}

\maketitle

\begin{abstract}
Vision transformers have excelled in various computer vision tasks but mostly rely on rigid input sampling using a fixed-size grid of patches. It limits their applicability in real-world problems, such as active visual exploration, where patches have various scales and positions.
Our paper addresses this limitation by formalizing the concept of input elasticity for vision transformers and introducing an evaluation protocol for measuring this elasticity.
Moreover, we propose modifications to the transformer architecture and training regime, which increase its elasticity.
Through extensive experimentation, we spotlight opportunities and challenges associated with such architecture.
\end{abstract}


\section{Introduction}

Vision Transformers (ViT)~\cite{dosovitskiy2020vit} achieve state-of-the-art results in many computer vision applications~\cite{carion2020endtoend,yu2022coca,kirillov2023segany}. They cut the image into a regular grid of non-overlapping patches and pass them as tokens to attention modules that exchange information between them. Many variations of the algorithm were created~\cite{chen2021asvit,rao2021dynamicvit,tang2022patch,yin2022avit}, including cross modal~\cite{akbari2021vatt} and long-term memory~\cite{wu2022memvit} solutions.

Nevertheless, almost all transformer-based methods assume that input tokens form a regular grid of $16 \times 16$ pixel patches, limiting their applicability in real-life problems. For example, in Active Visual Exploration (AVE)~\cite{pardyl2023active}, in which an agent has to navigate an environment by actively choosing successive observations while having a restricted field of view. AVE is commonly encountered in robotic applications, where an agent must localize dangers as quickly as possible based on incomplete images and observations of various scales and positions.

\begin{figure}
    \centering
    \begin{small}
    \begin{sc}
    \begin{tikzpicture}[node distance=0cm, auto]
    \tikzstyle{header} = [text centered]
    \tikzstyle{pict} = [inner sep=4px]
    \tikzstyle{arrow} = [thick,->,-{Latex[scale=1]}]
    
    \node [header, right] (selection) {\makecell[c]{Sampling\\selection}};
    \node [header, right=1.4cm of selection] (patches) {\makecell[c]{Extracted\\patches}};

    \node [pict, below=of selection, label={[align=right]left:Grid\\sampling}] (ssel) {\includegraphics[width=1.8cm]{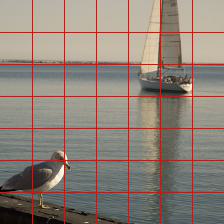}};
    \node [pict, below=of ssel, label={[align=right]left:Elastic\\sampling}] (esel) {\includegraphics[width=1.8cm]{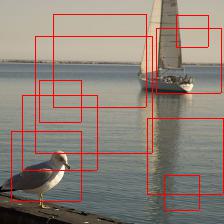}};

    \node [pict] at (ssel -| patches) (sx) {\includegraphics[width=1.8cm]{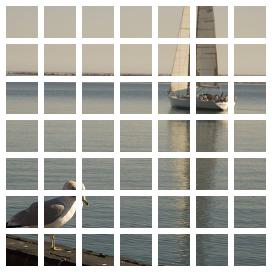}};
    \node [pict] at (esel -| patches) (ex) {\includegraphics[width=1.8cm]{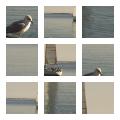}};

    \draw [arrow] (ssel) -- (sx);
    \draw [arrow] (esel) -- (ex);
    \end{tikzpicture}
    \end{sc}
    \end{small}
    \afterplot
    \caption{\textbf{Grid vs. elastic sampling:} patch sampling with arbitrary patch positions and scales creates new possibilities for more effective and efficient vision transformers.}
    \label{fig:teaser}
    \vspace{-0.1in}
\end{figure}

In this paper, we will consider two research questions. First, we ask if the standard ViT architectures are resilient to input perturbations that commonly occur in the vision tasks of embodied AI, such as AVE. For this purpose, we introduce an evaluation protocol that measures three necessary types of input elasticity a good model should showcase: \textit{scale elasticity}, \textit{position elasticity}, and \textit{missing data elasticity}.
We use this protocol to measure the input elasticity of common ViT architectures. Then, we move on to the second question: How can ViT input elasticity be increased by modifying model architecture and a training policy? We propose architectural and training modifications, named \textit{ElasticViT}\footnote{Our code is available at: \url{https://github.com/apardyl/beyondgrids}}, to increase the elasticity.

\newcommand{\imsize}{1.25cm}

\begin{figure*}[ht]
    \centering
    \begin{small}
    \begin{sc}
    \begin{tikzpicture}[node distance=.1cm, auto]
    
    \tikzstyle{header} = [text centered]
    \tikzstyle{pict} = [inner sep=0px]
    \tikzstyle{inputoutput} = [rectangle, rounded corners,text centered, draw=black]
    \tikzstyle{arrow} = [thick,->,-{Latex[scale=1]}]
    \tikzstyle{net} = [rectangle, rounded corners, minimum width=0cm, minimum height=1.8cm,text centered, draw=black, fill=blue!20, shape border rotate=270, trapezium angle=80]

    \node [pict, label={[align=right,rotate=90,label distance=-.5cm,yshift=.15cm]left:50\%}] (scale1) {\includesvg[width=\imsize]{figures/elasticities/scale/zoom0.5.svg}};
    \node [pict, below=of scale1, label={[align=right,rotate=90,label distance=-.5cm,yshift=.15cm]left:75\%}] (scale2) {\includesvg[width=\imsize]{figures/elasticities/scale/zoom0.75.svg}};
    \node [pict, below=of scale2, label={[align=right,rotate=90,label distance=-.5cm,yshift=.15cm]left:150\%}] (scale3) {\includesvg[width=\imsize]{figures/elasticities/scale/zoom1.5.svg}};

    \node [header, above=of scale1] (headerscale) {\makecell[c]{Scale}};

    \node [pict, right=.5cm of scale1, label={[align=right,rotate=90,label distance=-.5cm,yshift=.15cm]left:12.5\%}] (pos1) {\includesvg[width=\imsize]{figures/elasticities/position/shake12.5.svg}};
    \node [pict, label={[align=right,rotate=90,label distance=-.5cm,yshift=.15cm]left:25\%}] at (scale2 -| pos1) (pos2) {\includesvg[width=\imsize]{figures/elasticities/position/shake25.svg}};
    \node [pict, label={[align=right,rotate=90,label distance=-.5cm,yshift=.15cm]left:50\%}] at (scale3 -| pos1) (pos3) {\includesvg[width=\imsize]{figures/elasticities/position/shake50.svg}};

    \node [header] at (headerscale -| pos1) (headerposition) {\makecell[c]{Position}};

    \node [pict, right=.5cm of pos1, label={[align=right,rotate=90,label distance=-.5cm,yshift=.15cm]left:20\%}] (drop1) {\includesvg[width=\imsize]{figures/elasticities/dropout/drop20.svg}};
    \node [pict, label={[align=right,rotate=90,label distance=-.5cm,yshift=.15cm]left:40\%}] at (scale2 -| drop1) (drop2) {\includesvg[width=\imsize]{figures/elasticities/dropout/drop40.svg}};
    \node [pict, label={[align=right,rotate=90,label distance=-.5cm,yshift=.15cm]left:80\%}] at (scale3 -| drop1) (drop3) {\includesvg[width=\imsize]{figures/elasticities/dropout/drop80.svg}};

    \node [header] at (headerscale -| drop1) (headerdropout) {\makecell[c]{Missing data}};

    \node[draw,label=above:Perturbations,fit=(headerscale) (scale3) (scale3) (drop3), fit margins={left=.2cm,right=.2cm,bottom=.1cm,top=0cm}, rounded corners] (augs) {};

    \node [inputoutput, left=.5cm of augs] (input) {\includegraphics[width=1.5cm]{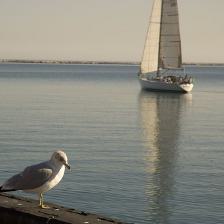}};
    \node [header] at (headerscale -| input) (headerinput) {\makecell[c]{Input\\image\\($I$)}};

    \node [inputoutput, right=.5cm of augs] (output) {\makecell[c]{\includegraphics[width=2cm]{figures/teaser/4.jpg}\\$+ \text{pos.enc.}$}};    
    \node [header] at (headerscale -| output) (headeroutput) {\makecell[c]{Patches\\($P'$)}};

    \draw [arrow] (input) -- (augs);
    \draw [arrow] (augs) -- (output);

    \node [net, right=.5cm of output] (vit) {\makecell[c]{Vision\\Transformer}};
    \draw [arrow] (output) -- (vit);
    \node [inputoutput, right=.5cm of vit] (pres) {$\pred(I)$};
    \draw [arrow] (vit) -- (pres);
        
    \end{tikzpicture}
    \end{sc}
    \end{small}
    \caption{\textbf{Evaluation protocol:} To analyze scale, position, and missing data elasticity we introduce three types of perturbations that are applied separately to each patch from the sampling grid.}
    \label{fig:eval-protocol}
\end{figure*}

We hope this work will draw the attention of the machine learning community to the important topic of input sampling strategies. It is crucial because, according to recent studies~\cite{beyer2023flexivit}, accommodating alternative patch resolutions can significantly impact the algorithm's real-life performance. We can summarize our contributions as follows:
\begin{itemize}
    \item We formalize the notion of input elasticity for vision transformers and provide a comprehensive evaluation protocol to assess it.
    \item We propose modifications to the vision transformer architecture and training policy, including a novel \textit{PatchMix} augmentation, that increase elasticity.
    \item We show that elastic sampling strategies can boost transformers' performance, especially for significantly limited input data.
\end{itemize}

\section{Related Work} \label{sec:relwork}

\textbf{Vision Transformers (ViTs).}
Vision transformer introduced in~\cite{dosovitskiy2020vit} is a versatile method due to its state-of-the-art performance obtained for many visual tasks~\cite{touvron2022deit}. Swin~\cite{liu2021Swin} modifies the attention mechanism to increase the locality of information exchange between tokens. AViT~\cite{chen2021asvit} demonstrates how transformer neural networks can be dynamically scaled. Pyramid ViT~\cite{wang2021pyramid} implements a multi-scale feature extraction model using ViT, inspired by previous work on convolutional neural networks. Vision Longformer~\cite{zhang2021multi} uses a token-based input format that enables seamless addition of global context tokens and control of the computation complexity. Other positional encoding methods for vision transformers were studied in~\cite{chu2022conditional,ronen2023vision,ke2021musiq}.

\textbf{ViT sampling strategies.}
Many works explore the possibility of using different grid resolutions as ViT inputs. Some perform different grid scale sampling during training~\cite{li2021efficient,wu2020multigrid}, and others introduce position and patch encoding rescaling tricks~\cite{beyer2023flexivit}. They usually improve ViT's accuracy across grids with varying resolutions and constant patches or when using native resolution grid sampling with a variable number of patches in a batch~\cite{dehghani2024patch}.

\textbf{ViTs applications.}
Initially, vision transformers were used in classical computer vision tasks, like object detection~\cite{carion2020endtoend} and semantic segmentation~\cite{kirillov2023segany}. However, as the field matures, the models are being deployed in real-world scenarios~\cite{li2021mvt}, where it is often necessary to understand input to the model as a collection of images captured from multiple views, each contributing partial information about a scene rather than a single image. Active visual exploration~\cite{Jha2023WACV,pardyl2023active} is one possible setup of such a real-world scenario. While performing the tasks, robots or drones often capture images that cover only part of the scene and rarely come in grid-aligned format with consistent scale. Therefore, it is crucial to provide higher input elasticity. Some of them were already considered, like ViT resiliency to missing data \cite{he2022masked} has already been conducted~\cite{tang2022patch,patchdropout23,yin2022avit}, but their cumulative effect on performance is unknown.

\section{Evaluation protocol} \label{sec:evalprotocol}

In this section, we first define three types of model elasticity corresponding to configurations that occur in embodied robotics data, and then we propose the evaluation protocol, which we implement by applying perturbations to the input sampling procedure. We use the protocol to rank the models by their overall elasticity.

\subsection{Elasticities} \label{eval_definitions}

We define model \textit{elasticity} as resilience to particular forms of input data configurations that can occur in real-life applications.

\textbf{Scale elasticity} is defined as resilience to the relative scale change across image tokens processed by the model.
Standard ViT algorithm uses fixed patch size and normalizes the size of each input image, while recent works allow for the use of native image resolution~\cite{dehghani2024patch} or different sampling grid sizes~\cite{beyer2023flexivit}.
We generalize the concept of scale elasticity to include varying scales of patches sampled from a single image. We measure it by sampling each patch individually and resampling it to the input token's standard $16 \times 16$ size. Note that this implementation might introduce \textit{missing data} to the input as shown in Fig.~\ref{fig:eval-protocol}.

\textbf{Positional elasticity} constitutes the resilience to positional change in input patches. Vanilla ViT is always trained with a rigid grid sampling even though, except for standard ViT positional encoding, the architecture does not require this.
We measure the positional elasticity by sampling image patches at the dense pixel-level resolution in contrast to the usual coarse grid-level resolution. Note that this sampling procedure might introduce \textit{missing data} as in Fig.~\ref{fig:eval-protocol}.

\textbf{Missing data elasticity} represents the resilience to missing image parts at the input.
Regular ViT assumes complete knowledge of the image, while many derivative works mask or remove a portion of input tokens~\cite{he2022masked,rao2021dynamicvit}.
In this work, we extend the concept of missing data to take into account situations where the tokens create an overlap of their receptive fields and do not cover the whole image area at the same time. Such overlapping might occur after patch scales or position perturbations, as shown in Fig.~\ref{fig:eval-protocol}. However, to measure the individual impact of missing data on model performance, independent of position and scale change, we drop a subset of grid-aligned patches from the input.

\begin{figure}
    \centering
    \begin{small}
    \begin{sc}
    \begin{tikzpicture}[node distance=0cm, auto]
    
    \tikzstyle{header} = [text centered]
    \tikzstyle{pict} = [inner sep=4px]
    \tikzstyle{arrow} = [thick,->,-{Latex[scale=1]}]

    \node [pict, label={[align=center]above:Image A}] (A) {\includegraphics[width=1.2cm]{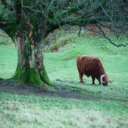}};
    \node [pict, right=of A, label={[align=center]above:Image B}] (B) {\includegraphics[width=1.2cm]{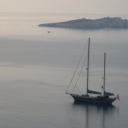}};
    \node [pict, right=of B, label={[align=center]above:CutMix}] (cut) {\includegraphics[width=1.45cm]{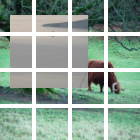}};
    \node [pict, right=of cut, label={[align=center]above:TokenMix}] (token) {\includegraphics[width=1.45cm]{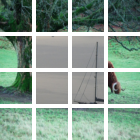}};
    \node [pict, right=of token, label={[align=center]above:PatchMix}] (patch) {\includegraphics[width=1.45cm]{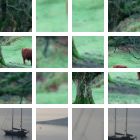}};

    \end{tikzpicture}
    \end{sc}
    \end{small}
    \vspace{-0.1in}
    \caption{\textbf{PatchMix:} We introduce PatchMix, an adaptation of CutMix and TokenMix to elasticity oriented training regime. PatchMix takes full advantage of the ElasticViT position and scale encoding, mixing randomly sampled patches.}
    \label{fig:patchmix}
\end{figure}

\subsection{Protocol}

The evaluation pipeline aims to assess the model elasticity by analyzing its resistance to changes in input sampling strategies. For this purpose, we create a set of patches based on the input image and then process this set through three perturbation functions corresponding to three considered elasticities. We use the perturbed set as a transformer input and report its performance. We provide conclusions on model elasticity by analyzing the performance obtained for different types of perturbations.

To strictly define the evaluation pipeline, let us consider image $I$ for which we generate set $P$ of patches $p = (x, y, s)$, where $x$ and $y$ denote the top-left corner's coordinates, and $s$ represents the relative scale (i.e. for a native patch of resolution $r \times r$, we sample a patch $r \cdot s \times r \cdot s$ and rescale it bilinearly to size $r \times r$). Initially, the coordinates $x$ and $y$ are from the regular grid and $s=1$. However, in the next step, we perturb them with three functions (presented in Fig.~\ref{fig:eval-protocol}) corresponding to the considered elasticities:
\begin{itemize}
\item $\E_{\scale(s_1, s_2)}(P)$ - introduces the scale perturbations,  sampling the $s$ parameter of every patch $p \in P$ independently and uniformly from range $[s_1, s_2]$.
\item $\E_{\pos(q)}(P)$ - applies positional perturbation, modifying $x$ and $y$ parameters of each patch $p \in P$, independently moving them by offsets sampled uniformly from range $[-r \cdot q, r \cdot q]$, where $r$ is the size of the patch.
\item $\E_{\miss(d)}(P)$ - adds missing data perturbations, dropping out $d$ patches from $P$ randomly with equal probability. 
\end{itemize}
Mathematically, this process can be described as follows:
\begin{equation*}
P' = \E_{\miss(d)}\circ \E_{\pos(q)}\circ \E_{\scale(s_1,s_2)}(P).
\end{equation*}
A disturbed set of patches $P'$ is used as an input of the transformer to test its elasticity. Elasticity is high if the predictions for $P'$ are as good as those obtained for $P$.

\section{Elastic ViT} \label{sec:ours}

In this section, we introduce a modification of the vision transformer architecture~\cite{dosovitskiy2020vit} for elastic input.



\subsection{Position and scale encoding} \label{exp:pos_scale_encodings}

Standard ViT~\cite{dosovitskiy2020vit} implementations utilize learnable positional embeddings. This is feasible because the grid sampling limits the number of possible patch positions. Unfortunately, such embeddings are not attainable with variable position and scale sampling. Each patch can be sampled at an arbitrary pixel position and with an arbitrary scale in elastic sampling. Therefore, the number of possible positions to be encoded is significantly larger. Consequently, we used a four-dimensional modification of the sine-cosine positional encoding of the original transformer model ~\cite{vaswani2017attention}. Let us recall the one-dimensional version of the sine-cosine encoding ($l$ is the length of the embedding and $i$ the $i$-th element of it): 
\begin{align*}
    \PE_{(\pos,2i)} &= \sin\left(\text{pos} / 10000^{2i/l}\right) \\
    \PE_{(\pos,2i+1)} &= \cos\left(\text{pos} / 10000^{2i/l}\right).
\end{align*}
To encode the position of a patch $p = (x, y, s)$, we separately encode the pixel coordinates of the patch upper-left and lower-right corner ($x, y, x + r \cdot s, y + r \cdot s$) and concatenate resultant embedding vectors.

Finally, we modify the input of a ViT to accept a set of cropped patches and a list of patch coordinates instead of a full image. The coordinates are used to generate positional encoding embeddings for each patch, allowing for continuous positional space. 

\subsection{Augmented training}

\newcommand{\splotwidth}{4.75cm}
\begin{figure*}[t]
\centering
\begin{subfigure}[b]{0.35\textwidth}
\begin{center}
\begin{small}
\begin{sc}
\begin{center}
\begin{tikzpicture}
\begin{axis}[
    title={Patch Scale},
    xlabel={Scale},
    ylabel={Accuracy (\%)},
    xtick={-8,0,8},
    xticklabels={0.5,1,2},
    xmin=-8,
    xmax=8,
    ymin=55,
    ymajorgrids=true,
    grid style=dashed,
    scale only axis,
    height=2.5cm,
    width=\splotwidth,
    legend style={at={(1.7,1.6)},anchor=north,/tikz/every even column/.append style={column sep=0.2cm}},
    legend cell align={center},
    legend columns=-1
]   
    \addplot[color=red]
    coordinates {
    (-8,74.200)(-7,76.928)(-6,79.364)(-5,81.156)(-4,82.268)(-3,82.760)(-2, 83.206)(-1, 83.264)(0,83.800)(1, 83.358)(2, 82.936)(3, 82.142)(4, 82.138)(5, 79.860)(6, 78.000)(7, 76.148)(8, 73.790)
    };
    \addlegendentry{ViT}

    \addplot[color=darkorange25512714]
    coordinates {
    (-8,73.1160022332764)(-7,75.5160023339844)(-6,77.3680022940064)(-5,79.3360023104858)(-4,80.7480022698975)(-3,81.8140023361206)(-2,82.6520023492432)(-1,83.2080022515869)(0,83.6380022329712)(1,83.3940023297119)(2,82.7840023770142)(3,81.7260022879028)(4,80.4420024073792)(5,78.8460023187256)(6,76.2080024163818)(7,72.784002151947)(8,67.6940020773316)
    };
    \addlegendentry{MAE}

    \addplot[color=forestgreen4416044]
    coordinates {
    (-8,44.3920014324951)(-7,52.0040017056274)(-6,58.6980018432617)(-5,66.3200020761108)(-4,71.7220022833252)(-3,75.5400023120117)(-2,79.3040024160767)(-1,82.038002303772)(0,83.7920022860718)(1,81.9900023025513)(2,80.7500022860718)(3,79.0120023202515)(4,76.7640022634888)(5,74.8220023034668)(6,72.0040021818543)(7,68.4400022116089)(8,62.9700019551086)
    };
    \addlegendentry{PVT}

    \addplot[color=gray]
    coordinates {
    (-8,54.3440017181397)(-7,63.5080020332336)(-6,68.1620022052002)(-5,74.1300022384644)(-4,78.0040023028565)(-3,80.0960023324585)(-2,81.806002359314)(-1,82.6700023193359)(0,83.4820023419189)(1,82.1840023556519)(2,81.9940023840332)(3,80.8060022799683)(4,78.2080023492432)(5,78.2840023413086)(6,76.106002303772)(7,72.7760022932434)(8,68.3600020809937)
    };
    \addlegendentry{Swin}

    \addplot[color=blue]
    coordinates {
    (-8,80.552)(-7,81.186)(-6,81.422)(-5,81.684)(-4,81.896)(-3,81.974)(-2, 82.016)(-1, 81.944)(0,82.036)(1, 81.780)(2, 81.240)(3, 80.790)(4, 79.742)(5, 79.594)(6, 78.898)(7, 78.244)(8, 77.504)
    };
    \addlegendentry{\ours{} (ours)}
\end{axis}
\end{tikzpicture}
\end{center}
\end{sc}
\end{small}
\end{center}
\afterplot
\caption{}
\label{fig:zoom-sample}
\end{subfigure}
\begin{subfigure}[b]{0.32\textwidth}
\begin{center}
\begin{small}
\begin{sc}
\begin{center}
\begin{tikzpicture}
\begin{axis}[
    title={Missing data},
    xlabel={Patches dropped (\%)},
    xtick={1,3,5,7,9},
    xticklabels={0, 20, 40, 60, 80},
    xmin=1,
    xmax=9,
    ymin=58,
    ymajorgrids=true,
    grid style=dashed,
    scale only axis,
    height=2.5cm,
    width=\splotwidth,
]
    \addplot[color=red]
    coordinates {
    (1,83.800)(2,83.316)(3,82.740)(4,81.852)(5,80.752)(6,79.042)(7,75.826)(8,70.154)(9,58.416)
    };
    \addlegendentry{ViT}
    
    \addplot[color=darkorange25512714]
    coordinates {
    (1,83.6380022329712)(2,83.3940022570801)(3,83.1200022940064)(4,82.5400023422241)(5,81.744002392273)(6,80.3920023129273)(7,78.2200022744751)(8,74.3040023487854)(9,66.8620021594238)
    };
    \addlegendentry{MAE}

    \addplot[color=forestgreen4416044]
    coordinates {
    (1,83.7920022860718)(2,82.6960022814942)(3,80.8140022793579)(4,78.5000023495483)(5,75.2080023130798)(6,70.9740022155762)(7,65.4080021047974)(8,58.7620018357849)(9,50.0320015971375)
    };
    \addlegendentry{PVT}

    \addplot[color=gray]
    coordinates {
    (1,83.4820023419189)(2,83.0380022857666)(3,82.3420022689819)(4,81.5560022836304)(5,80.2200023428345)(6,78.194002281189)(7,75.6340022589111)(8,71.0440022796631)(9,62.3500020339966)
    };
    \addlegendentry{Swin}
    
    \addplot[color=blue]
    coordinates {
    (1,82.036)(2,81.65)(3,81.276)(4,80.738)(5,80.046)(6,78.904)(7,77.328)(8,74.620)(9,68.582)
    };
    \addlegendentry{\ours{} (ours)}
    \legend{};
\end{axis}
\end{tikzpicture}
\end{center}
\end{sc}
\end{small}
\end{center}
\afterplot
\caption{}
\label{fig:dropout-sample}
\end{subfigure}
\begin{subfigure}[b]{0.32\textwidth}
\begin{center}
\begin{small}
\begin{sc}
\begin{center}
\begin{tikzpicture}
\begin{axis}[
    title={Patch position},
    xlabel={Patch shake (\%)},
    xtick={1,3,5,7,9},
    xticklabels={0, 12.5, 25.0, 37.5, 50},
    xmin=1,
    xmax=9,
    ymin=77,
    ymajorgrids=true,
    grid style=dashed,
    scale only axis,
    height=2.5cm,
    width=\splotwidth,
]
    \addplot[color=red]
    coordinates {
    (1,83.800)(2,83.728)(3,83.524)(4,83.222)(5,82.954)(6,82.504)(7,81.964)(8,81.532)(9,80.838)
    };
    \addlegendentry{ViT}

    \addplot[color=darkorange25512714]
    coordinates {
    (1,83.6380022329712)(2,83.6540022796631)(3,83.5340022680664)(4,83.3500022714233)(5,83.0480023071289)(6,82.7980022903442)(7,82.4940022140503)(8,82.1240023007202)(9,81.7680023339844)
    };
    \addlegendentry{MAE}

    \addplot[color=forestgreen4416044]
    coordinates {
    (1,83.7960022872925)(2,82.9260023580933)(3,81.7320023175049)(4,80.1100023526001)(5,78.2520023229981)(6,76.4020022477722)(7,74.1960022163391)(8,71.7180021903992)(9,69.016002142334)
    };
    \addlegendentry{PVT}

    \addplot[color=gray]
    coordinates {
    (1,83.4820023419189)(2,83.2860023699951)(3,83.0220023693848)(4,82.7500023706055)(5,82.2060024127197)(6,81.688002366333)(7,81.0780023422241)(8,80.3400023358154)(9,79.4460023458862)
    };
    \addlegendentry{Swin}
    
    \addplot[color=blue]
    coordinates {
    (1,82.036)(2,81.946)(3,82.012)(4,81.950)(5,81.830)(6,81.738)(7,81.704)(8,81.496)(9,81.294)
    };
    \addlegendentry{\ours{} (ours)}
    \legend{};
\end{axis}
\end{tikzpicture}
\end{center}
\end{sc}
\end{small}
\end{center}
\afterplot
\caption{}
\label{fig:shake-sample}
\end{subfigure}
\afterplot
\caption{\textbf{Isolated perturbations:} \textbf{(a)} The impact of changing the patch scale on accuracy. \ours{} can extract information from patches of various scales, being superior to other models in situations when information is lost either because patches do not cover the whole image or because the patches are of low resolution. \textbf{(b)} The effect of random patch dropout on the accuracy. \ours{} is more resilient to significant patch dropout than the other models and even outperforms the image reconstruction model (MAE) at the end of the measured spectrum. \textbf{(c)} The effect of applying positional perturbation on the accuracy. \ours{} can naturally interpret a whole range of possible position values. Being almost immune to the movement of sampled patches it eventually outperforms all but MAE for large perturbations. Note that the X-axis represents the percentage of patch movement relative to the patch size}
\end{figure*}

We propose a training regime modification for greater elasticity. As our baseline, we use the augmentation regime introduced in~\cite{touvron2022deit}, as it allows training the model on the ImageNet-1k~\cite{ILSVRC15} dataset without using any additional pre-training. We denote a ViT model trained with our modified regime as \textit{ElasticViT} in the following sections. A standard model trained without introduced elasticity is denoted as \textit{ViT} for comparison.

First, we introduce the elasticity functions into the augmentation pipeline. During training we use $\E_{\scale(1/3, 1)}$ (random patch sampling size in range $16$ to $48$), $\E_{\miss(0)}$ (no patch dropout), $\E_{\pos(\infty)}$ (unrestricted random patch positions). The native patch resolution (i.e., the resolutions all patches are rescaled to after sampling) is set to $16\times16$, and the native (scale = 1) image size is $224\times224$. We use $448\times448$ image resolution in the augmentation pipeline to accommodate for variable scale when required.

Second, we observe that the CutMix~\cite{yun2019cutmix} augmentation used in~\cite{touvron2022deit} is not optimal with the input elasticity we introduce. As our sampling strategy may change the proportions of images mixed by CutMix due to patch overlap, we must recalculate the mixed labels after applying elasticity functions to match the actual proportions of mixed images. Therefore, we replace it with PatchMix, a custom but comparable augmentation dedicated to ElasticViT. Similarly to TokenMix~\cite{liu2022tokenmix}, it mixes patches after sampling. However, contrary to TokenMix it is not dependant on standard grid sampling, and can take full advantage of the models ability to process arbitrary sampled patches (see Fig.~\ref{fig:patchmix} and supplementary materials). Moreover, we modify the MixUp~\cite{zhang2017mixup} augmentation, performing it after applying perturbations to patch sampling and ensuring that both sequences of patches to be mixed have the same patch scales element-wise.

\subsection{Experimental setup} \label{sec:setup}

We conduct experiments on a variety of state-of-the-art models with comparable parameter counts, differing in architecture or training policy. Namely, we evaluate our \ours, ViT from DeiT~III~\cite{touvron2022deit}, and MAE~\cite{he2022masked}, all based on ViT-B architecture, and further Swin-B~\cite{liu2021Swin}, and PVT-v2-B5~\cite{wang2021pvtv2}. All models were trained on ImageNet-1k~\cite{ILSVRC15}.

As the tested models significantly differ in architecture, we provide the implementation details of the missing data and position perturbations. The scale perturbation is implemented by patch resampling, independent of the model.

Because \ours, MAE, and DeiT~III model build on original~\cite{dosovitskiy2020vit} ViT architecture, we implement the action of $\E_{\miss}$ by removing the tokens from input entirely. Contrary to Swin and PVT, where attention mechanisms and positional encodings are more involved. For those models, we implement the action of $\E_{\miss}$ as zero-value masking of input tokens.

In all experiments, the magnitude of patch displacement is less than half of the patch size. Therefore, because Swin and PVT use relative positional encoding and $\E_{\miss}$ is implemented via patch masking, we implement the action of $\E_{\pos}$ by stitching sampled and masked patches into an image of the original size. Each patch is aligned to the nearest corresponding position in the grid.
ViT from DeiT~III and MAE use learned discrete-valued absolute positional encoding. Therefore, we encode the sampled patch position with an embedding corresponding to the position nearest to the patch.
\ours{} accepts continuous-valued positions, therefore we provide the model with true patch coordinates.

The experiments were performed on ImageNet-1k classification task. We report our results with both \textit{classification accuracy} and the \textit{number of tokens} used to achieve given results.
All training experiments were run using $4\times$NVIDIA A100 GPUs. \ours{} was trained for 800 epochs using the same training parameters as in~\cite{touvron2022deit}. For other models we used checkpoints provided by their authors.

To analyses transfer learning capabilities, we fine-tuned the above models for the MS COCO 2014~\cite{lin2014microsoft}, Pascal VOC 2007~\cite{everingham2009pascal} and ColonCancer~\cite{sirinukunwattana2016locality} datasets for $25$ epochs. Only the final fully-connected classification head was trained using standard grid sampling as in standard ViT. The training regime from~\cite{touvron2022deit} was used, excluding the utilization of MixUp and CutMix augmentations. We report the mean class average precision scores.

\newcommand{\mplotwidth}{4.75cm}
\begin{figure*}[th]
\centering
\begin{subfigure}[b]{0.35\textwidth}
\begin{center}
\begin{small}
\begin{sc}
\begin{center}
\begin{tikzpicture}
\pgfplotsset{
    scale only axis,
    height=2.5cm,
    width=\mplotwidth,
    ymajorgrids=true,
    grid style=dashed,
}
\begin{axis}[
    title={Patch position \& missing data},
    xlabel={Shake (\%)},
    ylabel={Accuracy (\%)},
    xtick={1,3,5,7,9},
    xticklabels={0, 12.5, 25.0, 37.5, 50},
    xmin=1,
    xmax=9,
    ymin=50,
    ymax=85,
    legend style={at={(1.7,1.6)},anchor=north,/tikz/every even column/.append style={column sep=0.2cm}},
    legend cell align={center},
    legend columns=-1
]
    \addplot[color=red]
    coordinates {
    (1,83.800)(2,83.228)(3,82.394)(4,81.304)(5,79.640)(6,77.354)(7,73.526)(8,66.772)(9,54.400)
    };
    \addlegendentry{ViT}

    \addplot[color=darkorange25512714]
    coordinates {
    (1,83.6380022329712)(2,83.348002277832)(3,82.7560023294067)(4,82.0280023730469)(5,80.9240023544312)(6,79.3100023754883)(7,77.0140022625732)(8,72.7700023068237)(9,65.0440019897461)
    };
    \addlegendentry{MAE}

    \addplot[color=forestgreen4416044]
    coordinates {
    (1,83.7940022885132)(2,81.8420022866821)(3,78.4780023065186)(4,74.1820022691345)(5,68.6700021708679)(6,61.9800019158936)(7,55.3060017726135)(8,48.8420015060425)(9,42.7080013845825)
    };
    \addlegendentry{PVT}

    \addplot[color=gray]
    coordinates {
    (1,83.4820023419189)(2,82.9040022961426)(3,82.0400023068237)(4,80.7600023428345)(5,78.9880023248291)(6,76.6420023468018)(7,73.4940022810364)(8,68.7240021282959)(9,59.8440018696594)
    };
    \addlegendentry{Swin}
    
    \addplot[color=blue]
    coordinates {
    (1,82.036)(2,81.576)(3,81.188)(4,80.726)(5,79.926)(6,78.720)(7,77.106)(8,74.200)(9,67.914)
    };
    \addlegendentry{\ours{} (ours)}
\end{axis}
\begin{axis}[
    axis x line* = bottom,
    axis y line = none,
    xlabel={Dropout (\%)},
    xtick={1,3,5,7,9},
    xticklabels={0, 20, 40, 60, 80},
    xmin=1,
    xmax=9,
]
    \addplot[opacity=0]
    coordinates {
    (1,1)(2,2) 
    };
    \pgfplotsset{every outer x axis line/.style={yshift=-1.cm}, every tick/.style={yshift=-1.cm}, every x tick label/.style={yshift=-1.cm} }
\end{axis}
\end{tikzpicture}
\end{center}
\end{sc}
\end{small}
\end{center}
\vspace{-.1in}
\caption{}
\label{fig:shake-dropout-sample}
\end{subfigure}
\begin{subfigure}[b]{0.32\textwidth}
\begin{center}
\begin{small}
\begin{sc}
\begin{center}
\begin{tikzpicture}
\pgfplotsset{
    scale only axis,
    height=2.5cm,
    width=\mplotwidth,
    ymajorgrids=true,
    grid style=dashed,
}
\begin{axis}[
    title={Patch position \& scale},
    xlabel={Shake (\%)},
    xtick={-8,0,8},
    xticklabels={50,0,50},
    xmin=-8,
    xmax=8,
    ymin=50,
    ymax=85,
]   
    \addplot[color=red]
    coordinates {
    (-8,65.886)(-7,71.750)(-6,76.144)(-5,79.218)(-4,81.152)(-3,82.474)(-2, 82.956)(-1, 83.258)(0,83.800)(1, 83.212)(2, 82.726)(3, 81.802)(4, 80.552)(5, 79.102)(6, 77.130)(7, 74.876)(8, 73.346)
    };
    \addlegendentry{ViT}

    \addplot[color=darkorange25512714]
    coordinates {
        (-8,71.7200022564697)(-7,73.9420022784424)(-6,76.3520022540283)(-5,78.5920023416138)(-4,80.3460023275757)(-3,81.6380022192383)(-2,82.6300023049927)(-1,83.182002315979)(0,83.6380022329712)(1,83.2960023364258)(2,82.7080023580933)(3,81.2500023510742)(4,79.7320023321533)(5,77.5800023658753)(6,74.486002263031)(7,69.7260021472168)(8,62.2660018537903)
    };
    \addlegendentry{MAE}

    \addplot[color=forestgreen4416044]
    coordinates {
    (-8,44.11000140625)(-7,51.6900016687012)(-6,58.5780017469788)(-5,65.1340020217896)(-4,70.3480021659851)(-3,75.0140022471619)(-2,78.760002272644)(-1,81.5680023220825)(0,83.7960022872925)(1,80.8940022921753)(2,77.9040023147583)(3,74.7180023164368)(4,70.7940021481323)(5,66.6460021466065)(6,61.7460019425964)(7,55.424001723175)(8,47.2240015296936)
    };
    \addlegendentry{PVT}

    \addplot[color=gray]
    coordinates {
    (-8,62.1960019395447)(-7,68.5020021328735)(-6,72.5140023477173)(-5,75.8700023162842)(-4,78.4980023703003)(-3,80.328002394104)(-2,81.5240023303223)(-1,82.4960023681641)(0,83.4820023419189)(1,81.9680023086548)(2,81.3400022479248)(3,79.6740023634338)(4,76.7480023127747)(5,76.002002252655)(6,72.6500021974182)(7,68.1140020976257)(8,61.7120019525147)
    };
    \addlegendentry{Swin}

    \addplot[color=blue]
    coordinates {
    (-8,79.754)(-7,80.668)(-6,81.074)(-5,81.414)(-4,81.670)(-3,81.828)(-2, 81.966)(-1, 81.954)(0,82.036)(1, 81.722)(2, 81.390)(3, 80.812)(4, 79.704)(5, 79.466)(6, 78.826)(7, 77.908)(8, 77.120)
    };
    \addlegendentry{\ours{} (ours)}
    \legend{};
\end{axis}
\begin{axis}[
    axis x line* = bottom,
    axis y line = none,
    xlabel={Scale},
    xtick={-8,0,8},
    xticklabels={0.5,1,2},
    xmin=-8,
    xmax=8,
]
    \addplot[opacity=0]
    coordinates {
    (1,1)(2,2) 
    };
    \pgfplotsset{every outer x axis line/.style={yshift=-1.cm}, every tick/.style={yshift=-1.cm}, every x tick label/.style={yshift=-1.cm} }
\end{axis}
\end{tikzpicture}
\end{center}
\end{sc}
\end{small}
\end{center}
\vspace{-.1in}
\caption{}
\label{fig:scale-shake-sample}
\end{subfigure}
\begin{subfigure}[b]{0.32\textwidth}
\begin{center}
\begin{small}
\begin{sc}
\begin{center}
\begin{tikzpicture}
\pgfplotsset{
    scale only axis,
    height=2.5cm,
    width=\mplotwidth,
    ymajorgrids=true,
    grid style=dashed,
}
\begin{axis}[
    title={Missing data \& scale},
    xlabel={Dropout (\%)},
    xtick={-8,0,8},
    xticklabels={80,0,80},
    xmin=-8,
    xmax=8,
    ymin=50,
    ymax=85,
]   
    \addplot[color=red]
    coordinates {
    (-8,37.488)(-7,53.104)(-6,65.354)(-5,73.266)(-4,77.896)(-3,80.532)(-2, 82.134)(-1, 82.768)(0,83.800)(1, 82.912)(2, 82.100)(3, 80.694)(4, 78.928)(5, 76.462)(6, 72.844)(7, 65.838)(8, 52.574)
    };
    \addlegendentry{ViT}

    \addplot[color=darkorange25512714]
    coordinates {
    (-8,64.1900019908142)(-7,71.1400023059082)(-6,74.8780023779297)(-5,77.6280023748779)(-4,79.5380024267578)(-3,81.0780023849487)(-2,82.1340023098755)(-1,82.8720023330689)(0,83.6380022329712)(1,82.9680023580933)(2,81.9360023382568)(3,79.8940023751831)(4,77.0640023796082)(5,72.300002308197)(6,64.5800020484924)(7,53.0040016287231)(8,35.7940011749268)
    };
    \addlegendentry{MAE}

    \addplot[color=forestgreen4416044]
    coordinates {
    (-8,38.2160012486267)(-7,41.0980012745667)(-6,46.7220014683533)(-5,54.7960017315674)(-4,63.2400019381714)(-3,69.8220022343445)(-2,76.3300023023987)(-1,80.888002350769)(0,83.7960022872925)(1,80.7500023339844)(2,77.7540022976685)(3,73.268002272644)(4,67.4980021461487)(5,59.9300019667053)(6,49.5360015444946)(7,37.3280012246704)(8,22.7880007736206)
    };
    \addlegendentry{PVT}

    \addplot[color=gray]
    coordinates {
    (-8,57.4540018063355)(-7,65.5660020921326)(-6,70.2260022299194)(-5,74.482002288208)(-4,77.576002286377)(-3,79.4380023156738)(-2,81.1180023413086)(-1,82.2780023034668)(0,83.4820023419189)(1,81.6540022769165)(2,80.54400236846920)(3,78.2480023277283)(4,73.5660021810913)(5,70.3440021522522)(6,62.8860019140625)(7,51.278001605835)(8,31.9400010548401)
    };
    \addlegendentry{Swin}
    
    \addplot[color=blue]
    coordinates {
    (-8,61.780)(-7,71.144)(-6,75.466)(-5,77.928)(-4,79.640)(-3,80.534)(-2, 81.202)(-1, 81.624)(0,82.036)(1, 81.312)(2, 80.808)(3, 79.730)(4, 78.176)(5, 77.468)(6, 75.468)(7, 72.648)(8, 67.170)
    };
    \addlegendentry{\ours{} (ours)}
    \legend{};
\end{axis}
\begin{axis}[
    axis x line* = bottom,
    axis y line = none,
    xlabel={Scale},
    xtick={-8,0,8},
    xticklabels={0.5,1,2},
    xmin=-8,
    xmax=8,
]
    \addplot[opacity=0]
    coordinates {
    (1,1)(2,2) 
    };
    \pgfplotsset{every outer x axis line/.style={yshift=-1.cm}, every tick/.style={yshift=-1.cm}, every x tick label/.style={yshift=-1.cm} }
\end{axis}
\end{tikzpicture}
\end{center}
\end{sc}
\end{small}
\end{center}
\vspace{-.1in}
\caption{}
\label{fig:scale-dropout-sample}
\end{subfigure}
\caption{\textbf{Combining perturbations:} The results of mixing multiple types of perturbations in one experiment. We observe that the performance of baseline models degrades quickly as perturbations add up. At the same time, the accuracy of ElasticViT remains stable for much longer, allowing more elastic input.}
\label{fig:mixes}
\afterplot
\end{figure*}

\section{Research questions} \label{research_questions}
Our goal was to investigate the resilience of transformers to input perturbations. For this purpose, we conducted extensive experiments, divided into six groups. The first three correspond to individual perturbations. The fourth group relates to their combinations. The fifth focuses on fundamental sampling strategies. In the sixth group, we examine a training trade-off between elasticity and base accuracy. Finally, in the last group we look into transferability of resilience to other datasets. All of them are described in the following subsections.

\subsection{Scale elasticity} \label{sec:scale-elasticity}


We maintain a consistent grid layout while introducing scale changes to each patch. The number of patches remains unchanged, but the perturbation inherently introduces overlapping or missing data to the input. This process is visually represented in Fig.~\ref{fig:eval-protocol} as \textit{Scale Elasticity}. The outcomes are illustrated in Fig.~\ref{fig:zoom-sample}.

\textbf{Is ViT robust with respect to scale?} \label{par:how-scale-elastic-stadard-vit}
All baseline models significantly decrease in accuracy when patch scale changes. Notably, the best-performing baseline model is the original supervised ViT while the worst are PVT and Swin. 


\textbf{Does applying randomized patch sampling in training improve scale elasticity at inference?}
Ours \ours{} despite having lower accuracy than other models at the base point ($82.04\%$ vs. $83.80\%$ of ViTs), due to high resilience to input scale variations, maintains its performance even at the ends of the measured spectrum and eventually outperforms all the other models.

\subsection{Missing data elasticity} \label{sec:missing-data}

Missing data can be simulated by adding perturbations to the input token set so part of the image is not represented by any of the cut patches. The patch scale experiment (Fig.~\ref{fig:zoom-sample}) introduced missing data aspects due to changing the scale of the patches, but it was a byproduct of other types of perturbations. We use a patch dropout scheme to isolate only the missing data aspect and call it the \textit{missing data} experiment.
The perturbation is depicted in Fig~\ref{fig:eval-protocol}. The results can be seen in Fig.~\ref{fig:dropout-sample}.

\textbf{How does input dropout affect ViT performance?}
Introducing dropout results in a consistent decline in the accuracy of all models, being increasingly noticeable as the percentage of dropped tokens increases. 
The best-performing baseline model is MAE, which is expected as it was trained on an image reconstruction task. The PVT significantly stands out from the other models, even from Swin, for which an identical dropout scheme was implemented and which has sparse attention too.

\begin{figure}
    \centering
    \begin{small}
    \begin{sc}
    \begin{tikzpicture}[node distance=0.2cm, auto]
    
    \tikzstyle{header} = [text centered]
    \tikzstyle{pict} = [inner xsep=4px]
    \tikzstyle{arrow} = [thick,->,-{Latex[scale=1]}]

    \node [pict, label={[align=center]right:EDGE\\sampling}] (sample) {\includegraphics[width=1.8cm]{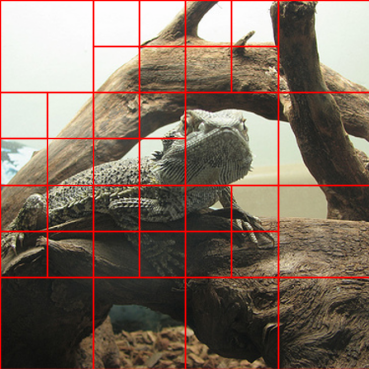}};

    \node [pict, above=of sample, label={[align=center]right:CENTRAL\\sampling}] (central) {\includegraphics[width=1.8cm]{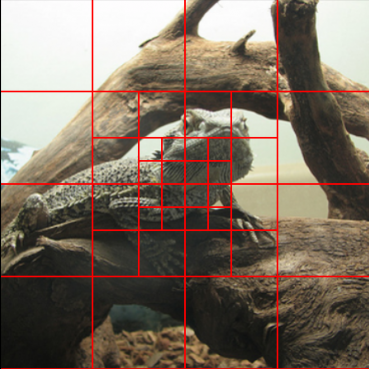}};

    \node[fit=(sample) (central) (central) (sample), fit margins={left=0cm,right=0cm,bottom=0cm,top=0cm}] (bbox) {};

    \node [pict, left=1cm of bbox, label={[align=center]left:Input}] (input) {\includegraphics[width=1.8cm]{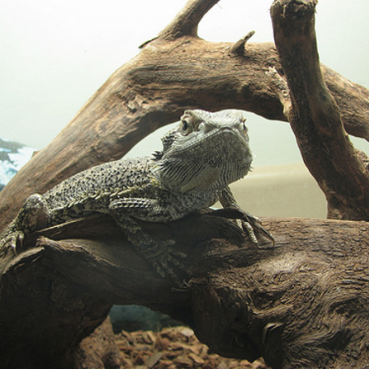}};

    \draw [arrow] (input) -- (sample);
    \draw [arrow] (input) -- (central);

    \end{tikzpicture}
    \end{sc}
    \end{small}
    \caption{\textbf{Patch sampling strategies:} We introduce two sampling strategies to show the benefits of elastic sampling. CENTRAL employ human retina-like higher density sampling in the center, EDGE samples based on weights provided by an edge detector (the more edges the denser sampling).}
    \label{fig:adaptive}
\end{figure}

\textbf{Do elastic perturbations during training enhance performance when dealing with missing data?}
ElasticViT exhibits a comparable behavior to the baselines when it comes to missing data, displaying a decline in performance as the token count decreases. Nevertheless, the rate of accuracy loss is gentler. Notably, ElasticViT surpasses MAE performance after the removal of 75\% of tokens.
This is significant because in the training the \ours{} always accepted the full token count, contrary to MAE, which was trained to reconstruct the image from the 25\% of input tokens.

\subsection{Positional elasticity} \label{sec:pos-elasticity}

Elastic vision transformers should not be limited to accepting only rigid sets of patch positions. In \textit{patch position} experiment, we introduce a concept called "patch shake" – a randomized shift in the position of the original patches from a grid by a specified percentage of the patch size. ElasticViT's positional embeddings readily adapt to these positional changes. However, for the other models, we had to modify their positional embeddings to make this experiment viable. We encode each patch with embedding corresponding to the grid layout position being nearest to the patch. The patch shake operation is illustrated in Fig.~\ref{fig:eval-protocol}. The outcomes can be observed in Fig.~\ref{fig:shake-sample}.

\textbf{Is ViT elastic to positional perturbations?}
Patch shake results in a minor accuracy reduction for all the models except PVT. It's important to note that patch shake can inherently introduce missing data and overlap perturbations. Therefore, this experiment doesn't isolate which specific perturbation has the most substantial impact on performance. Nonetheless, it's worth considering that the classification task might be inherently position agnostic, as ImageNet classification could be efficiently accomplished by treating patches as "bags of words" without explicit positional context.

\begin{figure}[t]
\begin{center}
\begin{small}
\begin{sc}
\begin{center}
\begin{tikzpicture}
\begin{axis}[
    title={Sampling strategies},
    xlabel={No. input patches (log scale)},
    ylabel={Accuracy (\%)},
    xtick={16,64,196,784},
    xmin=16,
    xmax=784,
    ymin=38,
    ymax=87,
    ymajorgrids=true,
    grid style=dashed,
    scale only axis,
    height=3cm,
    width=.8\linewidth,
    legend pos=south east,
    legend cell align={left},
    xmode=log,
    log ticks with fixed point,
]   
    \addplot[color=red, dashed]
    coordinates {
    (16, 19.798)(25, 44.942)(36,60.908)(49, 73.590) (64, 75.938) (81, 79.002) (100, 79.756) (121, 80.950) (144, 81.812) (169, 82.550) (196, 83.800) (225, 83.136) (256, 83.304) (324, 82.742) (361, 82.454) (441, 81.750) (529, 80.516) (625, 79.704) (784, 77.438)
    };
    \addlegendentry{ViT, grid}
    
    \addplot[color=red]
    coordinates {
    (16, 19.806)(24, 45.478)(28, 62.018)(40, 66.946)(49,73.594)(54,75.102)(76,79.440)(124,82.498)(196,83.8)(208,83.502)(244, 83.508)(302, 81.634)(388, 80.404)(496, 79.954)(628, 78.576)(784, 77.43)
    };
    \addlegendentry{ViT central}
    
    \addplot[color=blue, dashed]
    coordinates {
    (16, 40.566)(25, 58.076)(36,67.428)(49, 73.160) (64, 75.408) (81, 77.794) (100, 78.510) (121, 79.710) (144, 80.606) (169, 81.340) (196, 82.036) (225, 82.276) (256, 82.400) (324, 82.752) (361, 82.818) (441, 83.032) (529, 83.072) (625, 83.176) (784, 83.072)
    };
    \addlegendentry{\ours{}, grid}

    \addplot[color=blue]
    coordinates {
    (16, 40.558)(24, 59.78)(28, 65.062)(40, 70.29)(49,73.164)(54,74.766)(76,78.498)(124,80.912)(196,82.034)(208,82.248)(244, 82.492)(302, 82.650)(388, 82.794)(496, 82.910)(628, 83.072)(784, 83.072)
    };
    \addlegendentry{\ours{}, central}

    \addplot[color=forestgreen4416044]
    coordinates {
    (16,52.6)(25,64.5)(36,70.35)(49,74.57)(64,77.13)(81,78.55)(100,79.65)(121,80.21)(144,80.9)(169,81.16)(196,81.4)
    (225,81.6)(256,81.9)(324,81.93)(361,82.1)(441,82.12)(529,82.33)(625,82.39)(784,83.072)};
    \addlegendentry{\ours{}, edge}
\end{axis}
\end{tikzpicture}
\end{center}
\end{sc}
\end{small}
\end{center}
\afterplot
\caption{\textbf{Central and adaptive patch sampling:} Results of applying two patch sampling strategies (CENTRAL and EDGE) compared to standard grid sampling at inference. We observe that by leveraging input elasticity we can gain a significant improvement in accuracy when the number of sampled patches is limited.}
\label{fig:central-sample}
\end{figure}

\textbf{Does training with randomized sampling improve positional elasticity?}
Like the missing data experiment (Sec.~\ref{sec:missing-data}), ElasticViT exhibits greater resilience to patch shake than the baselines. ElasticViT outperforms all models but MAE when the patch positions are shaken by approximately 40\% of the patch size.
Notably, the accuracy of \ours{} remained almost constant with respect to the increase of patch shake, while the second-best model in this regard (MAE) lost 2 p.p. of accuracy. 

\subsection{Combining perturbations} \label{sec:mix-elasticity}
We conduct a series of tests by combining the previously introduced input perturbations to assess their collective impact on performance. The scaling ranges for perturbation parameters are the same as in the earlier experiments. The objective is to determine whether ElasticViT's resilience will continue to outperform that of the original ViT in more complex scenarios.

\textbf{How does positional elasticity combine with missing data elasticity?}
In \textit{patch position \& missing data} experiment, we introduce a combination of dropout and shake perturbations, and the results are depicted in Fig.~\ref{fig:shake-dropout-sample}.
Notably, the detrimental effect on ViT's performance closely aligns with the summed impact of shake and dropout when performed independently.
This lets \ours{} surpass MAE at the 60\% dropout threshold, instead of 75\% as in the \textit{missing data}~\ref{fig:dropout-sample} experiment.

\begin{figure}[t]
\begin{center}
\begin{small}
\begin{sc}
\begin{center}
\begin{tikzpicture}
\pgfplotsset{
    scale only axis,
    height=3cm,
    width=.8\linewidth,
    ymajorgrids=true,
    grid style=dashed,
}
\begin{axis}[
    title={Training trade-off},
    xlabel={Scale (\%)},
    ylabel={Accuracy (\%)},
    xtick={-8,0,8},
    xticklabels={0.5,1,2},
    xmin=-8,
    xmax=8,
    ymin=35,
    ymax=85,
    legend style={at={(0.5,0.03)},anchor=south},
    legend cell align={left}
]   
    \addplot[color=red]
    coordinates {
    (-8,43.76200142)(-7,57.97400179)(-6,66.92400213)(-5,72.60600219)(-4,76.68600228)(-3,79.46600241)(-2,81.38600239)(-1,82.58800238)(0,83.37600232)(1,82.18000231)(2,80.66400224)(3,78.36800228)(4,75.26600226)(5,71.42600216)(6,65.44800208)(7,54.59200179)(8,37.26400119)
    };
    \addlegendentry{0\% \ours{}}

    \addplot[color=darkorange25512714]
    coordinates {
    (-8,71.52000222)(-7,76.12800232)(-6,78.78800238)(-5,80.01800238)(-4,81.10600235)(-3,81.65000233)(-2,82.2280024)(-1,82.55000232)(0,83.09400234)(1,81.57200237)(2,81.53200228)(3,79.62400231)(4,77.40400232)(5,73.92200224)(6,68.94200213)(7,60.40600187)(8,47.92400146)
    };
    \addlegendentry{15\% \ours{}}

    \addplot[color=forestgreen4416044]
    coordinates {
    (-8,71.196)(-7,75.846)(-6,78.232)(-5,79.638)(-4,80.68)(-3,81.448)(-2,81.966)(-1,82.376)(0,82.972)(1,81.478)(2,81.29)(3,79.47)(4,77.32)(5,73.76)(6,68.908)(7,60.4)(8,48.208)
    };
    \addlegendentry{30\% \ours{}}

    \addplot[color=gray]
    coordinates {
    (-8,70.69000225)(-7,75.75800235)(-6,78.33200226)(-5,79.8040023)(-4,80.63600237)(-3,81.38000231)(-2,81.88600233)(-1,82.35200231)(0,82.9360023)(1,81.57000228)(2,81.12000233)(3,79.13000231)(4,76.46200231)(5,72.69400221)(6,67.16000199)(7,57.7880018)(8,44.4120014)
    };
    \addlegendentry{70\% \ours{}}
    
    \addplot[color=blue]
    coordinates {
    (-8,70.44000226)(-7,75.30200224)(-6,78.01200231)(-5,79.54800234)(-4,80.56800229)(-3,81.16400232)(-2,81.64800234)(-1,81.99400233)(0,82.00000227)(1,81.50600238)(2,80.90400225)(3,79.94200234)(4,78.17200234)(5,75.76400225)(6,71.71200216)(7,64.49000202)(8,52.28000171)
    };
    \addlegendentry{100\% \ours{}}
\end{axis}
\begin{axis}[
    axis x line* = bottom,
    axis y line = none,
    xlabel={Dropout (\%)},
    xtick={-8,0,8},
    xticklabels={80,0,80},
    xmin=-8,
    xmax=8,
]
    \addplot[opacity=0]
    coordinates {
    (1,1)(2,2) 
    };
    \pgfplotsset{every outer x axis line/.style={yshift=-1.2cm}, every tick/.style={yshift=-1.2cm}, every x tick label/.style={yshift=-1.2cm} }
\end{axis}
\begin{axis}[
    axis x line* = bottom,
    axis y line = none,
    xlabel={Shake (\%)},
    xtick={-8,0,8},
    xticklabels={50,0,50},
    xmin=-8,
    xmax=8,
]
    \pgfplotsset{every outer x axis line/.style={yshift=-2.4cm}, every tick/.style={yshift=-2.4cm}, every x tick label/.style={yshift=-2.4cm} }
    \addplot[opacity=0]
    coordinates {
    (1,1)(2,2) 
    };
\end{axis}
\end{tikzpicture}
\end{center}
\end{sc}
\end{small}
\end{center}
\afterplot
\caption{\textbf{Training for elasticity:} Evaluation of training regimes with different proportions of the grid and \ours{} augmented training (see Sec.~\ref{sec:ours}) in training batches. For validation we apply all three perturbations. Note that only a small number of \ours{} augmented training batches ($<15\%$) is required to improve elasticity significantly.}
\label{fig:training-tradeoff}
\end{figure}

\textbf{What is the interaction between positional elasticity and scale data elasticity?}
In \textit{patch position \& patch scale} experiment, we combine the zoom perturbation with the shake perturbation. The results are shown in Fig.~\ref{fig:scale-shake-sample}. The ratio of performance decline between ViT and ElasticViT remained consistent with that reported in the previous experiment for patches twice as large. However, as patches became smaller, the performance gap between both algorithms widened. In the most altered scenario, ElasticViT outperforms ViT with a nearly 20 percentage point lead.

\textbf{How does scale elasticity combine with missing data elasticity?}
In \textit{missing data \& patch scale} experiment, we combine dropout with patch scale change, and the results are illustrated in Fig.~\ref{fig:scale-dropout-sample}. The outcomes resemble those of the grid zoom experiment (Sec.~\ref{par:how-scale-elastic-stadard-vit}) but exhibit a more pronounced negative impact of the perturbations. ElasticViT begins to outperform ViT at approximately the midway point of the perturbation intensity scale.

\subsection{Patch sampling strategies}
\label{sec:multiscale-adaptation}
The elastic features demonstrated by ElasticViT create an opportunity to apply a multi-scale approach to partitioning images into patches of different scales in a non-uniform manner. Given the results obtained in the grid experiments, there's a possibility that this approach can lead to performance improvement without sacrificing accuracy compared to standard grid sampling (referred to as~\textit{GRID}).

\begin{figure}[t]
\begin{center}
\begin{small}
\begin{sc}
\begin{center}
\begin{tikzpicture}
\pgfplotsset{
    scale only axis,
    height=3cm,
    width=.8\linewidth,
    ymajorgrids=true,
    grid style=dashed,
}
\begin{axis}[
    title={MS COCO},
    xlabel={Scale (\%)},
    ylabel={mAP (\%)},
    xtick={-8,0,8},
    xticklabels={0.5,1,2},
    xmin=-8,
    xmax=8,
    ymin=28,
    ymax=60,
    legend style={at={(0.5,0.03)},anchor=south},
    legend cell align={left},
    legend columns=2
]   
    \addplot[color=red]
    coordinates {
    (-8,34.44661745)(-7,39.63408968)(-6,43.28053684)(-5,45.61417924)(-4,47.21231775)(-3,48.28758482)(-2,49.17981536)(-1,50.07333205)(0,50.92112853)(1,49.71823582)(2,49.02001526)(3,47.68255287)(4,45.83080034)(5,43.18475799)(6,39.39260364)(7,34.06346305)(8,28.23588199)
    };
    \addlegendentry{ViT}

    \addplot[color=darkorange25512714]
    coordinates {
        (-8,44.24650417)(-7,47.36735665)(-6,49.71502994)(-5,51.40705605)(-4,52.57568011)(-3,53.56881412)(-2,54.42355263)(-1,55.16253253)(0,55.94836693)(1,55.316971)(2,54.40296214)(3,53.02999704)(4,51.27642405)(5,48.65176587)(6,45.5633944)(7,41.64144616)(8,35.84748445)
    };
    \addlegendentry{MAE}

    \addplot[color=forestgreen4416044]
    coordinates {
    (-8,35.97219265)(-7,37.44335402)(-6,40.78890532)(-5,44.67439525)(-4,48.6131548)(-3,51.76395281)(-2,54.36011921)(-1,56.23343231)(0,58.10276963)(1,56.18552886)(2,54.18054598)(3,51.52717713)(4,48.06816668)(5,43.5101641)(6,38.1193615)(7,32.83912018)(8,26.82481085)
    };
    \addlegendentry{PVT}

    \addplot[color=gray]
    coordinates {
    (-8,39.25648722)(-7,44.34799931)(-6,47.3939814)(-5,50.04615733)(-4,51.71966412)(-3,53.37010641)(-2,54.47647199)(-1,55.42558809)(0,56.31964758)(1,55.4357384)(2,54.54413394)(3,52.96664625)(4,49.83839041)(5,48.68217851)(6,45.09830177)(7,39.62304992)(8,30.65232724)
    };
    \addlegendentry{Swin}

    \addplot[color=blue]
    coordinates {
    (-8,43.42022656)(-7,48.13718817)(-6,49.75104975)(-5,50.58388736)(-4,51.13968064)(-3,51.55767508)(-2,51.85234301)(-1,52.0102303)(0,52.15408588)(1,52.11819873)(2,51.96748911)(3,51.67225015)(4,51.08738975)(5,50.50936535)(6,49.32283905)(7,47.56690999)(8,43.43508238)
    };
    \addlegendentry{\ours{} (ours)}
\end{axis}
\begin{axis}[
    axis x line* = bottom,
    axis y line = none,
    xlabel={Dropout (\%)},
    xtick={-8,0,8},
    xticklabels={80,0,80},
    xmin=-8,
    xmax=8,
]
    \addplot[opacity=0]
    coordinates {
    (1,1)(2,2) 
    };
    \pgfplotsset{every outer x axis line/.style={yshift=-1.2cm}, every tick/.style={yshift=-1.2cm}, every x tick label/.style={yshift=-1.2cm} }
\end{axis}
\begin{axis}[
    axis x line* = bottom,
    axis y line = none,
    xlabel={Shake (\%)},
    xtick={-8,0,8},
    xticklabels={50,0,50},
    xmin=-8,
    xmax=8,
]
\pgfplotsset{every outer x axis line/.style={yshift=-2.4cm}, every tick/.style={yshift=-2.4cm}, every x tick label/.style={yshift=-2.4cm} }
    \addplot[opacity=0]
    coordinates {
    (1,1)(2,2) 
    };
\end{axis}
\end{tikzpicture}
\end{center}
\end{sc}
\end{small}
\end{center}
\afterplot
\caption{\textbf{Transfer learning:} Results of fine tuning the last layers of models for multi-label classification of the MS COCO 14 dataset. All models were trained with standard grid sampling. We observe, that for high perturbation rate \ours{} outperforms baselines.}
\label{fig:coco}
\end{figure}

First, we test a hypothesis that, for the ImageNet, the central portion of an image might carry more significance than its periphery. To explore this, we iteratively divide the patches closest to the center of the image into four smaller patches. We refer to this algorithm as \textit{CENTRAL}, see Fig.~\ref{fig:adaptive}. The rationale is that smaller patches offer a higher sampling resolution and introduce more data into the input, potentially leading to improved accuracy. The CENTRAL algorithm is designed so that the neutral point corresponds to a token count of 196, which aligns with the original ViT grid of $14\times14$ and a single scale. Lower values in this context result in larger patches near the image periphery, while higher values position smaller patches toward the center.

Next, we extend on the previous experiment, analyzing if an adaptive patch sampling strategy can improve the results, especially in low patch count scenarios. We create a toy adaptive sampling algorithm, which we refer to as \textit{EDGE}, based on the Canny edge detector~\cite{canny1986computational}. The method starts by computing a bitmap of edges with a large $\sigma$ value for the detector. The map is then divided into $112\times112$ patches, which are then sorted by their sums over the bitmap. Then, the algorithm iteratively divides a patch with the highest sum larger than the minimal size of $8\times8$, into 4 new patches, which are added to the list, preserving the order. The algorithm is repeated until the number of patches reaches the target. Visualization of EDGE is presented in Fig.~\ref{fig:adaptive}.

\textbf{Does using higher resolution of the grid in the center improve accuracy?}
Our findings for the CENTRAL algorithm are presented in Fig.~\ref{fig:central-sample}. In the case of the vanilla ViT model, we observed issues with resilience to scale changes. Depending on the non-standard scale content of the input dataset, ViT results can be either negatively or positively influenced by the CENTRAL algorithm. Conversely, when considering ElasticViT, the differences are significantly reduced, and there is a higher likelihood that CENTRAL sampling leads to improved model performance.

\textbf{Can adaptive patch size sampling perform better?}
The results of this experiment are presented in Fig.~\ref{fig:central-sample} for the \ours{} model. We observe, that EDGE performs significantly better than both GRID and CENTRAL algorithms in low patch count scenarios (less than 64 patches). When limited to 25 patches, it achieves over $6\%$ improvement compared to \ours{} and $20\%$ over standard ViT, increasing to $12\%$ and $32\%$ (accordingly) when only 16 patches can be sampled. Consequently, we claim that even very simple adaptive patch sampling methods can reduce the number of tokens needed to achieve targeted performance.

\subsection{Trade-offs} \label{sec:tradeoff}

This section explores a training trade-off for elastic sampling. Further experiments on trade-off in inference are provided in supplementary materials.


\textbf{What is the tradeoff between accuracy and elastic training?}
A consistent pattern emerges in all our experiments comparing ViT and ElasticViT: ViT generally exhibits higher accuracy than ElasticViT when evaluated under the original, unperturbed grid sampling scenario. However, as we introduce perturbations into the evaluation, ElasticViT can surpass ViT. Earlier we performed experiments that juxtapose classical single-scale grid training against a fully randomized sampling method of ElasticViT, representing two extremes in the spectrum.

Fig.~\ref{fig:training-tradeoff} illustrates the results of training with mixed simple grid sampling (standard ViT training) and a fully randomized setup (\ours{} training regime, see~\ref{sec:ours}). The evaluation was conducted by applying all three perturbation variants at the same time.

The findings reveal that for simple grid evaluation, elastic training has a relatively minor negative impact on accuracy. However, in the case of randomized evaluation that requires elasticity, even with only 15\% of the training data containing randomized patches, there is a substantial gain of more than ten percentage points in accuracy. This implies that in practical applications, fine-tuning the network can be accomplished using just a fraction of perturbed input data while having minimal repercussions on baseline (simple grid) accuracy. Yet, this approach offers increased resilience to perturbations, demonstrating the practical utility of elastic inputs in the training process.

\subsection{Transfer Learning}

We evaluate elasticity of models pre-trained on the ImageNet-1k dataset and fine-tuned to the target tasks.

\textbf{Does elasticity capabilities transfer to other datasets?}
\label{sec:transfer-learning}
An important question that remains to be answered is whether the elasticity capabilities observed on the ImageNet-1k dataset transfer to other datasets and tasks. We analyze this problem on the matter of multi-label classification of the MS COCO 2014 dataset as described in Sec.~\ref{sec:setup}. We perform evaluation applying all three perturbations (see Sec.~\ref{sec:evalprotocol}) at the same time. Results of this experiment are presented in Fig.~\ref{fig:coco}, PVT, Swin and MAE perform the best when no perturbation are introduced. However, \ours{} exhibits the best consistency in results across all perturbation range. The standard ViT model performs the worst, even without any perturbations, indicating worse generalization capabilities. Results for VOC and ColonCancer are provided in supplementary materials.

\section{Conclusions}
This study examines Vision Transformers (ViT) and their adaptability to varying input sampling strategies required in real-world applications. We introduce an evaluation protocol to assess ViT's resistance to input perturbations, including scale, missing data, and positional changes.

Standard ViT models prove to be vulnerable to these perturbations, displaying a significant performance drop due to potential overfitting during training. In contrast, the modified ViT we proposed exhibits better resilience thanks to randomized training, maintaining or improving performance in various scenarios, see overall comparison in Fig.~8 of the supplementary material.

Our experiments also explore adaptive patch sampling using CENTRAL and EDGE algorithms, which ElasticViT benefits from, particularly in scenarios with fewer patches. Adaptive patch sampling efficiently reduces token requirements while preserving target performance.

Future work in this area should focus on refining the adaptive patch sampling strategies and further investigating the trade-offs between downscaling input tokens and patch dropout under computational constraints. Additionally, research should aim to apply these findings to real-world applications, enhancing transformer models' adaptability in practical contexts.

\section*{Acknowledgments}

{\small
This paper has been supported by the Horizon Europe Programme (HORIZON-CL4-2022-HUMAN-02) under the project "ELIAS: European Lighthouse of AI for Sustainability", GA no. 101120237, and by National Science Centre, Poland (grant no. 2023/49/N/ST6/02465, 2022/47/B/ST6/03397, 2022/45/B/ST6/02817, and 2023/50/E/ST6/00469). Some experiments were performed on servers purchased with funds from a grant from the Priority Research Area (Artificial Intelligence Computing Center Core Facility) under the Strategic Programme Excellence Initiative at Jagiellonian University. We gratefully acknowledge Polish high-performance computing infrastructure PLGrid (HPC Center: ACK Cyfronet AGH) for providing computer facilities and support within computational grant no. PLG/2024/017483.
}

{\small
\bibliographystyle{ieee_fullname}
\bibliography{main}
}

\end{document}


\title{Supplementary material for\\Beyond Grids: Exploring Elastic Input Sampling for Vision Transformers}

\author{Adam Pardyl$^{1,2,3}$ \qquad Grzegorz Kurzejamski$^{1}$ \qquad Jan Olszewski$^{1,4}$ \\
Tomasz Trzciński$^{1,5,6}$ \qquad Bartosz Zieliński$^{1,2}$ \\
$^1$IDEAS NCBR \\
$^2$Jagiellonian University, Faculty of Mathematics and Computer Science \\
$^3$ Jagiellonian University, Doctoral School of Exact and Natural
Sciences \\
$^4$ University of Warsaw \qquad $^5$ Warsaw University of Technology \qquad $^6$ Tooploox
}

\maketitle

\section{Additional experiments}

\subsection{PatchMix ablation}

\begin{figure}
    \centering
    \includegraphics[width=.95\linewidth]{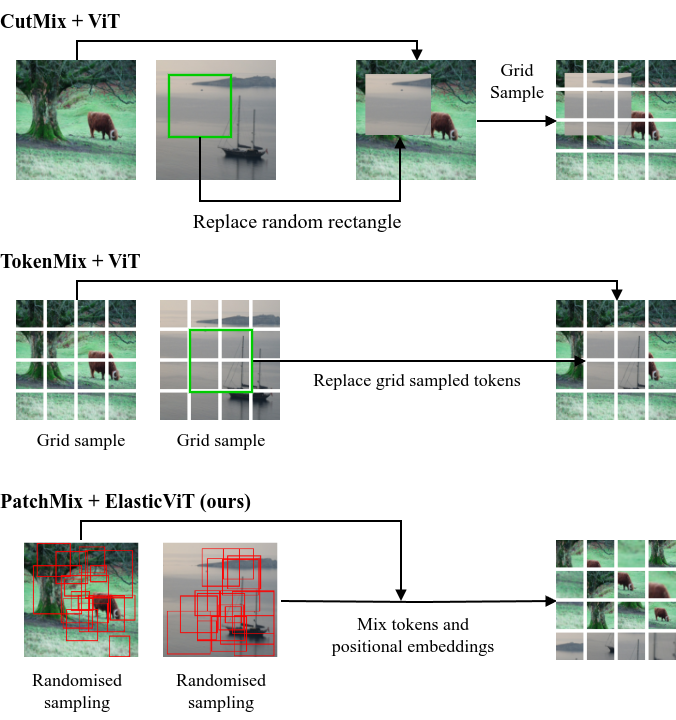}
    \caption{\textbf{PatchMix:} Standard CutMix poses an issue in ElasticViT training regime, as proportions of mixed targets may change in randomized sampling. TokenMix replaces patches after sampling, incidentally eliminating this issue, but nevertheless relies on grid sampling. Our PatchMix takes full advantage of the ElasticViT position and scale encoding, and enables mixing of randomly sampled patches of different scales.}
    \label{fig:patchmix}
\end{figure}

To evaluate the effectiveness of PatchMix (see Fig.~\ref{fig:patchmix} for visualization) we perform an ablation study, training the model with the regime presented in Sec. 4 of the main paper, but without the PatchMix augmentation applied. The results are presented in Tab.~\ref{tab:patchmix_ablation}. The PatchMix augmentation improves the accuracy of ElasticViT by over 1.5\% on ImageNet-1k.

\begin{table}[t]
    \centering
    \begin{tabular}{rl}
    \toprule
        ElasticViT & Accuracy \\
        \midrule
        without PatchMix & 80.67\% \\
        with PatchMix & 82.04\% \\
        \bottomrule
    \end{tabular}
    \caption{\textbf{PatchMix ablation study:} The effectiveness of the PatchMix augmentation evaluated on the Imagnet-1k dataset. The test was performed without any introduced perturbations. We observe that PachMix provides over 1.5\% gain in accuracy.}
    \label{tab:patchmix_ablation}
\end{table}

\subsection{Grid density}

Continuing on scale elasticity experiments presented in Sec.~5.1, we decided to investigate the resistance of ViT architectures to change of grid density. In real-world scenarios, scale changes are quite common. However, when conducting synthetic experiments with ViT, the typical approach is to maintain a consistent input image resolution and grid layout density.

\begin{figure}
    \centering
    \begin{small}
    \begin{sc}
    \begin{tikzpicture}[node distance=0cm, auto]
    
    \tikzstyle{header} = [text centered]
    \tikzstyle{pict} = [inner sep=4px]
    \tikzstyle{arrow} = [thick,->,-{Latex[scale=1]}]

    \node [pict, label={[align=center]above:\\$0.5$}, label={[align=left]left:Grid density}] (scale1) {\includesvg[width=1.5cm]{figures/elasticities/grid/gridx0.5.svg}};
    \node [pict, right=of scale1, label={[align=center]above:Scale\\$1$}] (scale2) {\includesvg[width=1.5cm]{figures/elasticities/grid/gridx1.svg}};
    \node [pict, right=of scale2, label={[align=center]above:\\$1.33$}] (scale3) {\includesvg[width=1.5cm]{figures/elasticities/grid/gridx1.33.svg}};

    \node [pict, below=of scale1, label={[align=left]left:Grid zoom}] (zoom1) {\includesvg[width=1.5cm]{figures/elasticities/scale/zoom0.5.svg}};
    \node [pict, right=of zoom1] (zoom2) {\includesvg[width=1.5cm]{figures/elasticities/scale/zoomx1.svg}};
    \node [pict, right=of zoom2] (zoom3) {\includesvg[width=1.5cm]{figures/elasticities/scale/zoomx1.33.svg}};


        
    \end{tikzpicture}
    \end{sc}
    \end{small}

    \caption{\textbf{Scale elasticities:} We use two perturbation scenarios, that change the size of a patch in a grid. The first (grid density) changes the number of patches. The second (grid zoom) keeps same number of patches but changes theirs size.}
    \label{fig:gridscalezoom}
\end{figure}

In this setup, we adjust the grid density, thereby altering the number of patches while using the same input image. This modification process is illustrated in Fig.~\ref{fig:gridscalezoom} as \textit{Grid density} and compared to the \textit{Grid zoom} perturbation shown in Sec.~5.1. The outcomes for standard ViT, MAE and \ours{} are presented in Fig.~\ref{fig:grid-sample}. PVT and Swin models were omitted in this evaluation, as changing the grid density in those models is not trivial due their internal structure. We observe, that all models perform similarly well when decreasing the density of the grid, while only \ours{} can utilize denser sampling to its advantage.

\subsection{Is it better to down-scale input or dropout patches?}

The results from our previous experiments naturally lead to an important question: When faced with computational constraints, is it more beneficial to remove an entire patch from the input or to rescale neighboring patches in order to maintain complete image coverage, even at the cost of changing the token scale? To investigate this, we conduct experiments where we iteratively select a random $2\times2$ patch block from the uniform $14\times14$ patch grid and modify it in two ways. The first is to replace a $2\times2$ block of patches with a single larger patch, preserving the same coverage. In the second, we remove three out of four patches within the block. In both cases, these operations reduced the input token set by three, either through a dropout operation or by changing the scale of the patches.

 \begin{figure}[t]
\begin{center}
\begin{small}
\begin{sc}
\begin{center}
\begin{tikzpicture}
\begin{axis}[
    title={Grid density},
    xlabel={Scale},
    ylabel={Accuracy (\%)},
    xtick={-8,0,8},
    xticklabels={0.5,1,2},
    xmin=-8,
    xmax=8,
    ymajorgrids=true,
    grid style=dashed,
    scale only axis,
    height=4cm,
    width=.8\linewidth,
    legend pos=south west,
    legend cell align={left}
]   
    \addplot[color=red]
    coordinates {
    (-8,77.438)(-7,79.704)(-6,80.516)(-5,81.750)(-4,82.454)(-3,82.742)(-2, 83.304)(-1, 83.136)(0,83.800)(1, 82.550)(2, 81.812)(3, 80.950)(4, 79.756)(5, 79.002)(6, 77.092)(7, 75.938)(8, 73.590)
    };
    \addlegendentry{ViT}
    
    \addplot[color=darkorange25512714]
    coordinates {
    (-8,68.9940021536255)(-7,74.4660022700501)(-6,76.3540024267578)(-5,78.7780024365234)(-4,80.3300026123047)(-3,81.3500022039795)(-2,82.9700023007202)(-1,83.1620023236084)(0,83.6380022329712)(1,82.1360023852539)(2,81.4280023931885)(3,80.7260024014282)(4,79.1820023461914)(5,78.9820022982788)(6,76.8900023596192)(7,75.3840023373413)(8,73.454002270813)
    };
    \addlegendentry{MAE}
    
    \addplot[color=blue]
    coordinates {
    (-8,83.072)(-7,83.176)(-6,83.072)(-5,83.032)(-4,82.818)(-3,82.752)(-2, 82.400)(-1, 82.276)(0,82.036)(1, 81.340)(2, 80.606)(3, 79.710)(4, 78.510)(5, 77.794)(6, 75.802)(7, 75.408)(8, 73.160)
    };
    \addlegendentry{\ours{} (ours)}
\end{axis}
\end{tikzpicture}
\end{center}
\end{sc}
\end{small}
\end{center}
\afterplot
\caption{\textbf{Grid density}: The impact of changing the grid density (see Fig.~\ref{fig:gridscalezoom}) on accuracy. \ours{} can utilize extra information from denser grid sampling ($0.5$), outperforming the original ViT.}
\label{fig:grid-sample}
\end{figure}

The results of these experiments are depicted in Fig.~\ref{fig:drop-vs-rescale-sample}. Surprisingly, ElasticViT exhibits almost no discernible difference in performance between the dropout and rescale operations, with dropout slightly outperforming rescaling at the extremes. Again, This would suggest a potential overfit with results better when having only 25\% of the image covered by patches than having 100\% coverage but with two times lower sampling resolution. In contrast, for the original ViT model, there is a visible distinction in performance between the two methods of limiting token counts, with the rescaling option proving to be a much more effective solution at the extremes.

\begin{figure}[t]
\begin{center}
\begin{small}
\begin{sc}
\begin{center}
\begin{tikzpicture}
\begin{axis}[
    title={Dropout or rescale},
    xlabel={Replaced $2\times 2$ patch blocks (\%)},
    ylabel={Accuracy (\%)},
    xtick={1,...,6},
    xticklabels={0, 20, 40, 60, 80, 100},
    xmin=1,
    xmax=6,
    ymin=50,
    ymajorgrids=true,
    grid style=dashed,
    scale only axis,
    height=6cm,
    width=.8\linewidth,
    legend pos=south west,
    legend cell align={left}
]
    \addplot[color=red, dashed]
    coordinates {
    (1,83.800)(2,82.860)(3,81.724)(4,80.072)(5,77.622)(6,73.590)
    };
    \addlegendentry{ViT, rescale}
    
    \addplot[color=red]
    coordinates {
    (1,83.800)(2,83.010)(3,81.938)(4,79.654)(5,75.526)(6,67.388)
    };
    \addlegendentry{ViT, drop}

    \addplot[color=darkorange25512714, dashed]
    coordinates {
    (1,83.6380022329712)(2,82.9060023114014)(3,82.0000023077393)(4,80.4780023693848)(5,77.8980023309326)(6,73.454002270813)
    };
    \addlegendentry{MAE, rescale}

    \addplot[color=darkorange25512714]
    coordinates {
    (1,83.6380022329712)(2,83.0200022940064)(3,82.2900024008179)(4,80.750002315979)(5,77.804002364502)(6,73.0160022406006)
    };
    \addlegendentry{MAE, drop}

    \addplot[color=forestgreen4416044, dashed]
    coordinates {
    (1,83.7940022885132)(2,82.4280023147583)(3,79.1980022982788)(4,71.7960021466064)(5,53.7820016984558)(6,19.232000693512)
    };
    \addlegendentry{PVT, rescale}

    \addplot[color=forestgreen4416044]
    coordinates {
    (1,83.7960022872925)(2,82.6900023022461)(3,80.2360022940064)(4,75.8980022654724)(5,65.9820020574951)(6,39.1980012892151)
    };
    \addlegendentry{PVT, drop}

    \addplot[color=gray, dashed]
    coordinates {
    (1,83.4820023419189)(2,82.5680022976685)(3,80.9360023077393)(4,78.1740022943115)(5,73.6020022897339)(6,64.2580020178223)
    };
    \addlegendentry{Swin, rescale}

    \addplot[color=gray]
    coordinates {
    (1,83.4820023419189)(2,82.6340023529053)(3,81.2220023358154)(4,78.9640023199463)(5,75.260002288208)(6,69.0300021659851)
    };
    \addlegendentry{Swin, drop}

    \addplot[color=blue, dashed]
    coordinates {
    (1,82.036)(2,81.334)(3,80.596)(4,79.282)(5,77.216)(6,73.16)
    };
    \addlegendentry{\ours{}, rescale}
    \addplot[color=blue]
    coordinates {
    (1,82.036)(2,81.392)(3,80.522)(4,79.380)(5,77.364)(6,74.164)
    };
    \addlegendentry{\ours{}, drop}
\end{axis}
\end{tikzpicture}
\end{center}
\end{sc}
\end{small}
\end{center}
\afterplot
\caption{\textbf{Reducing number of patches:} Evaluation of the trade-off between lower resolution sampling and patch dropout. The X-axis represents the number of $2\times2$ patch blocks either replaced by a single lower-resolution patch or kept with only one patch while dropping the rest. We observe that the chosen strategy only affects ViT significantly at a high number of replacements, whereas ElasticViT remains unaffected due to its greater elasticity in handling missing data.}
\label{fig:drop-vs-rescale-sample}
\end{figure}

\subsection{Transfer learning (continued)}

\subsubsection{Pascal VOC dataset}

In this section we show additional results for transfer learning elasticity, evaluated on the Pascal VOC dataset. We follow the same training and evaluation setup as in Sec.~5.7 of the main paper. Results of the experiments are presented in Fig.~\ref{fig:voc}. We observer, that standard ViT performs the best for the native resolution, but is outperformed by \ours{} when elasticity is introduced. Both PVT and Swin performs slightly worse, and surprisingly, the self-supervised pre-trained MAE performs the worst, failing to achieve event half of the mean average precision score of ViT and \ours{}.

\begin{figure}[t]
\begin{center}
\begin{small}
\begin{sc}
\begin{center}
\begin{tikzpicture}
\pgfplotsset{
    scale only axis,
    height=4cm,
    width=.8\linewidth,
    ymajorgrids=true,
    grid style=dashed,
}
\begin{axis}[
    title={PASCAL VOC},
    xlabel={Scale (\%)},
    ylabel={mAP (\%)},
    xtick={-8,0,8},
    xticklabels={0.5,1,2},
    xmin=-8,
    xmax=8,
    ymin=10,
    ymax=90,
    legend style={at={(0.5,0.03)},anchor=south},
    legend cell align={left}
]   
    \addplot[color=red]
    coordinates {
    (-8,65.795)(-7,73.553)(-6,77.724)(-5,79.646)(-4,81.576)(-3,82.766)(-2,83.521)(-1,83.999)(0,84.666)(1,83.31)(2,82.931)(3,81.69)(4,80.141)(5,77.663)(6,72.955)(7,64.896)(8,53.516)
    };
    \addlegendentry{ViT}

    \addplot[color=darkorange25512714]
    coordinates {
        (-8,27.108)(-7,30.262)(-6,33.049)(-5,35.678)(-4,37.172)(-3,38.341)(-2,38.346)(-1,37.801)(0,38.276)(1,38.033)(2,36.703)(3,33.145)(4,29.692)(5,26.323)(6,24.049)(7,22.041)(8,19.564)
    };
    \addlegendentry{MAE}

    \addplot[color=forestgreen4416044]
    coordinates {
    (-8,34.084)(-7,41.797)(-6,49.394)(-5,56.109)(-4,63.467)(-3,68.314)(-2,72.743)(-1,76.557)(0,77.102)(1,75.191)(2,72.014)(3,67.615)(4,61.099)(5,53.605)(6,42.575)(7,32.789)(8,23.861)
    };
    \addlegendentry{PVT}

    \addplot[color=gray]
    coordinates {
    (-8,30.446)(-7,37.651)(-6,44.688)(-5,51.282)(-4,57.125)(-3,62.537)(-2,68.198)(-1,71.6)(0,71.766)(1,72.174)(2,67.358)(3,61.049)(4,48.943)(5,43.427)(6,37.857)(7,30.963)(8,23.734)
    };
    \addlegendentry{Swin}

    \addplot[color=blue]
    coordinates {
    (-8,73.575)(-7,78.015)(-6,79.365)(-5,80.214)(-4,81.112)(-3,81.317)(-2,81.578)(-1,81.946)(0,82.113)(1,82.048)(2,81.5)(3,81.305)(4,80.863)(5,80.093)(6,78.913)(7,77.053)(8,72.54)
    };
    \addlegendentry{\ours{} (ours)}
\end{axis}
\begin{axis}[
    axis x line* = bottom,
    axis y line = none,
    xlabel={Dropout (\%)},
    xtick={-8,0,8},
    xticklabels={80,0,80},
    xmin=-8,
    xmax=8,
]
    \addplot[opacity=0]
    coordinates {
    (1,1)(2,2) 
    };
    \pgfplotsset{every outer x axis line/.style={yshift=-1.2cm}, every tick/.style={yshift=-1.2cm}, every x tick label/.style={yshift=-1.2cm} }
\end{axis}
\begin{axis}[
    axis x line* = bottom,
    axis y line = none,
    xlabel={Shake (\%)},
    xtick={-8,0,8},
    xticklabels={50,0,50},
    xmin=-8,
    xmax=8,
]
\pgfplotsset{every outer x axis line/.style={yshift=-2.4cm}, every tick/.style={yshift=-2.4cm}, every x tick label/.style={yshift=-2.4cm} }
    \addplot[opacity=0]
    coordinates {
    (1,1)(2,2) 
    };
\end{axis}
\end{tikzpicture}
\end{center}
\end{sc}
\end{small}
\end{center}
\afterplot
\caption{\textbf{PASCAL VOC transfer learning}: Results of fine tuning the last layers of models for multi-label classification of the PASCAL VOC 2007 dataset. All models were trained with standard grid sampling. We observe, that for high perturbation rate \ours{} outperforms baselines. Note the surprisingly poor performance of the self-supervised pre-trained MAE.}
\label{fig:voc}
\end{figure}

\subsubsection{ColonCancer dataset}

We further test transfer learning properties of our model, evaluating it on the ColonCancer dataset. The dataset consists of histopathological images to be classified as malignant or benign. As previously, we train only the final linear layer of the model. For standard ViT we apply regular grid sampling, our ElasticViT uses EDGE sampling as described in the main paper. Results are presented in Tab.~\ref{tab:cc}, showing superior performance of ElasticViT. We attribute this superior performance to EDGE sampling, which extracts more patches in regions containing cell nuclei, enabling ElasticViT to create a better representation of the data.

\begin{table}[t]
    \centering
    \begin{tabular}{rcl}
    \toprule
        Model & Sampling type & Accuracy \\
        \midrule
        ViT & GRID & 82\% \\
        ElasticViT & EDGE & 87\% \\
        \bottomrule
    \end{tabular}
    \caption{\textbf{ColonCancer transfer learning:} Results of fine-tuning the last layers of models for binary classification of the ColonCancer dataset. The standard ViT model was run with grid sampling, while our ElasticViT model utilized EDGE sampling as described in the main paper. We observe that our model outperforms the standard ViT, benefiting from variable scale sampling.}
    \label{tab:cc}
\end{table}

\subsection{Stability of elastic evaluation}
\begin{figure}[t]
    \centering
    \includegraphics[scale=.5]{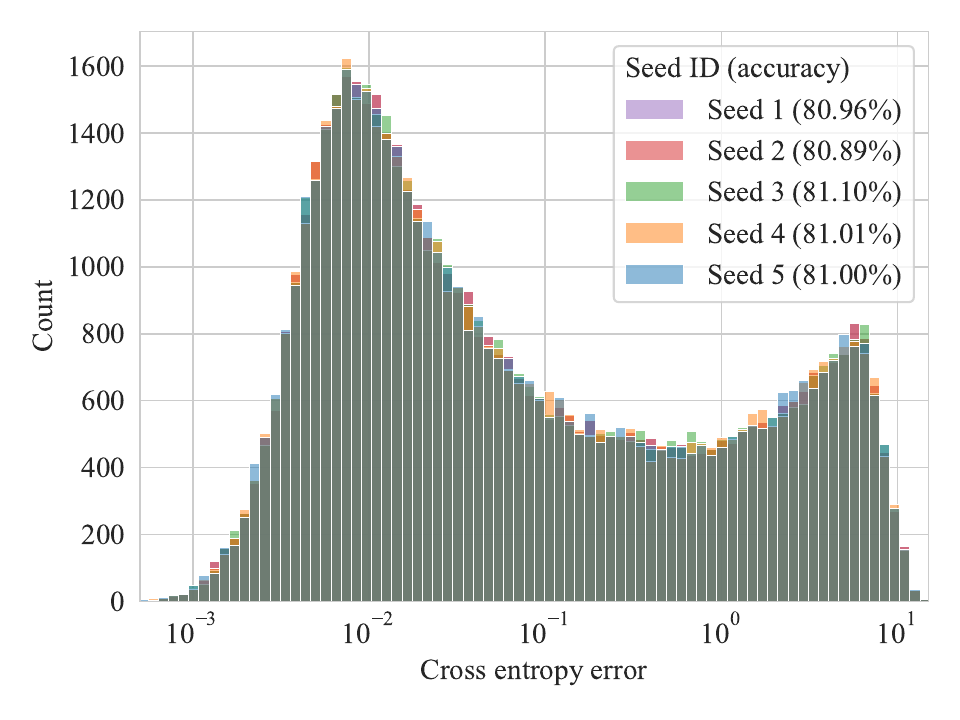}
    \caption{\textbf{Elastic sampling impact on performance across multiple runs:} The histogram represents the distribution of classification error for five inference runs with different random seeds used for elastic sampling. We observe that the differences between runs are insignificant. Note that the accuracy scores of particular runs are provided in the plot legend.}
    \label{fig:seedcheck}
\end{figure}

As our elasticity benchmark introduces randomness, we run the evaluation pipeline on the same trained model multiple times to check for any fluctuations in performance across different random seeds. The differences in performance are insignificant over multiple runs of elastic sampling, as shown in the histogram of performance standard deviations in Fig.~\ref{fig:seedcheck}. This consistency can be attributed to the large size of the ImageNet-1k validation set.

\subsection{Patch redundancy}

\begin{figure}[t]
\begin{center}
\begin{small}
\begin{sc}
\begin{center}
\begin{tikzpicture}
\begin{axis}[
    title={Patch redundancy},
    xlabel={Redundant patches},
    ylabel={Accuracy (\%)},
    xtick={0,98,196},
    xticklabels={0,+98,+196},
    xmin=0,
    xmax=196,
    ymin=81,
    ymax=84,
    ymajorgrids=true,
    grid style=dashed,
    scale only axis,
    height=4cm,
    width=.8\linewidth,
    legend pos=south west,
    legend cell align={left}
]   
    \addplot[color=red]
    coordinates {
    (0,83.800)(14,83.6300022894287)(28,83.5880023202515)(42,83.4080023330689)(56,83.3060022860718)(70,83.2960023062134)(84,83.1600023751831)(98,83.1160023620605)(112,82.9620022775269)(126,82.8240023101807)(140,82.8220023364258)(154,82.7120022756958)(168,82.6500023751831)(182,82.4460023348999)(196,82.6080023468018)
    };
    \addlegendentry{ViT}
    
    \addplot[color=darkorange25512714]
    coordinates {
    (0,83.6380022329712)(14,83.5440022390747)(28,83.4960022766113)(42,83.3000022998047)(56,83.3620022659302)(70,83.148002253418)(84,83.0060022091675)(98,82.9380022369385)(112,82.7960022888184)(126,82.6260022964478)(140,82.6300022943115)(154,82.4080023196411)(168,82.280002260437)(182,82.3020022433472)(196,82.066)
    };
    \addlegendentry{MAE}
    
    \addplot[color=blue]
    coordinates {
    (0,82.036)(14,82.0480023139954)(28,82.1640023265076)(42,82.1760022859192)(56,82.2980022731018)(70,82.3020022349548)(84,82.3480023153687)(98,82.3020023014832)(112,82.328002297821)(126,82.3640022996521)(140,82.4520023216248)(154,82.4480023033142)(168,82.3920022712708)(182,82.4140022583008)(196,82.3480022109985)
    };
    \addlegendentry{\ours{} (ours)}
\end{axis}
\end{tikzpicture}
\end{center}
\end{sc}
\end{small}
\end{center}
\afterplot
\caption{\textbf{Patch redundancy}: The impact of adding randomly sampled patches in addition of a standard sampling grid. We observe, that standard ViT and MAE lose performance, as those randomly sampled patches as treated as noise. In contrast, ElasticViT can utilize the extra information to improve the result.}
\label{fig:grid-redundancy}
\end{figure}

\begin{figure}[t]
\begin{center}
\begin{small}
\begin{sc}
\begin{center}
\begin{tikzpicture}[
  treatment line/.style={rounded corners=1.5pt, line cap=round, shorten >=1pt},
  treatment label/.style={font=\small},
  group line/.style={ultra thick},
]

\begin{axis}[
  clip={false},
  axis x line={center},
  axis y line={none},
  axis line style={-},
  xmin={1},
  ymax={0},
  scale only axis={true},
  width={.5\linewidth},
  ticklabel style={anchor=south, yshift=1.3*\pgfkeysvalueof{/pgfplots/major tick length}, font=\small},
  every tick/.style={draw=black},
  major tick style={yshift=.5*\pgfkeysvalueof{/pgfplots/major tick length}},
  minor tick style={yshift=.5*\pgfkeysvalueof{/pgfplots/minor tick length}},
  title style={yshift=\baselineskip},
  xmax={5},
  ymin={-3.5},
  height={4\baselineskip},
  xtick={1,2,3,4,5},
  minor x tick num={3},
  title={Average ranking based on performance},
]

\draw[treatment line] ([yshift=-2pt] axis cs:1.1428571428571428, 0) |- (axis cs:0.726190476190476, -2.5)
  node[treatment label, anchor=east] {elasticvit};
\draw[treatment line] ([yshift=-2pt] axis cs:2.5714285714285716, 0) |- (axis cs:0.726190476190476, -3.5)
  node[treatment label, anchor=east] {mae};
\draw[treatment line] ([yshift=-2pt] axis cs:2.7142857142857144, 0) |- (axis cs:5.416666666666667, -4.0)
  node[treatment label, anchor=west] {vit};
\draw[treatment line] ([yshift=-2pt] axis cs:3.5714285714285716, 0) |- (axis cs:5.416666666666667, -3.0)
  node[treatment label, anchor=west] {swin};
\draw[treatment line] ([yshift=-2pt] axis cs:5.0, 0) |- (axis cs:5.416666666666667, -2.0)
  node[treatment label, anchor=west] {pvt};
\draw[group line] (axis cs:2.5714285714285716, -2.3333333333333335) -- (axis cs:2.7142857142857144, -2.3333333333333335);

\end{axis}
\end{tikzpicture}
\end{center}
\end{sc}
\end{small}
\end{center}
\afterplot
\caption{\textbf{Critical difference diagram}: Average performance ranking of \ours{} and baseline methods for ImageNet with high perturbation (most extreme settings). \ours{} significantly outperforms remaining approaches, while more structured methods (PVT and Swin) perform significantly worse than simple ViT. Moreover, the difference between ViT and MAE is insignificant. Notice that this diagram was generated for rankings generated based on results from Fig.~4 and Fig.~5 of the main paper. 
}
\label{fig:critdist}
\end{figure}

\begin{figure*}
    \makebox[\textwidth][c]{\includegraphics[width=.95\textwidth]{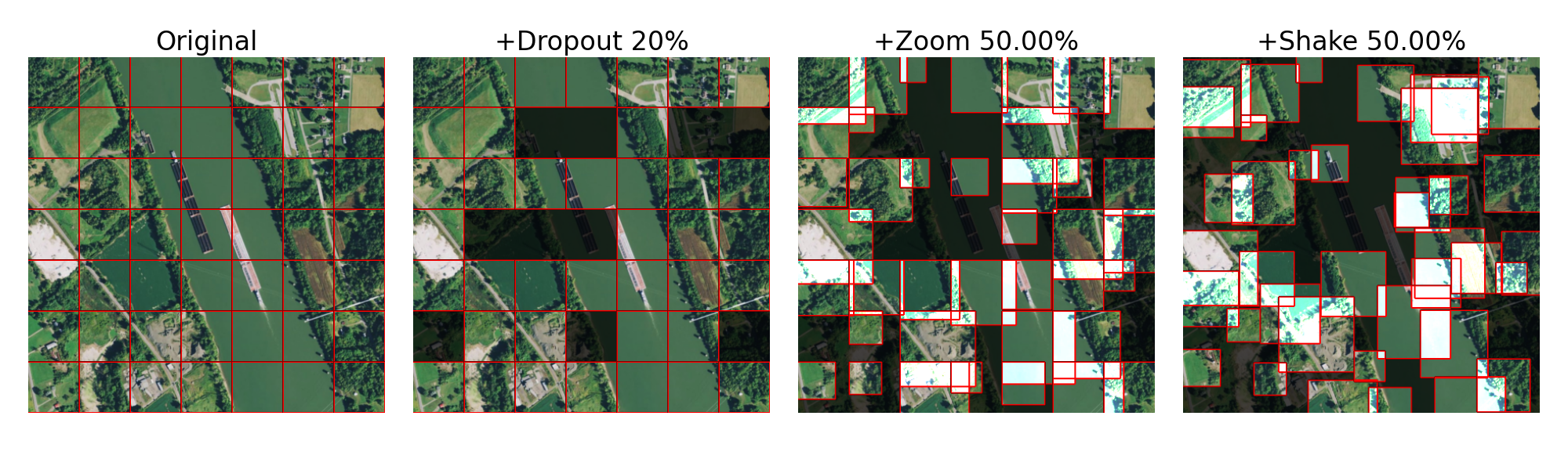}} \\ 
    \makebox[\textwidth][c]{\includegraphics[width=.95\textwidth]{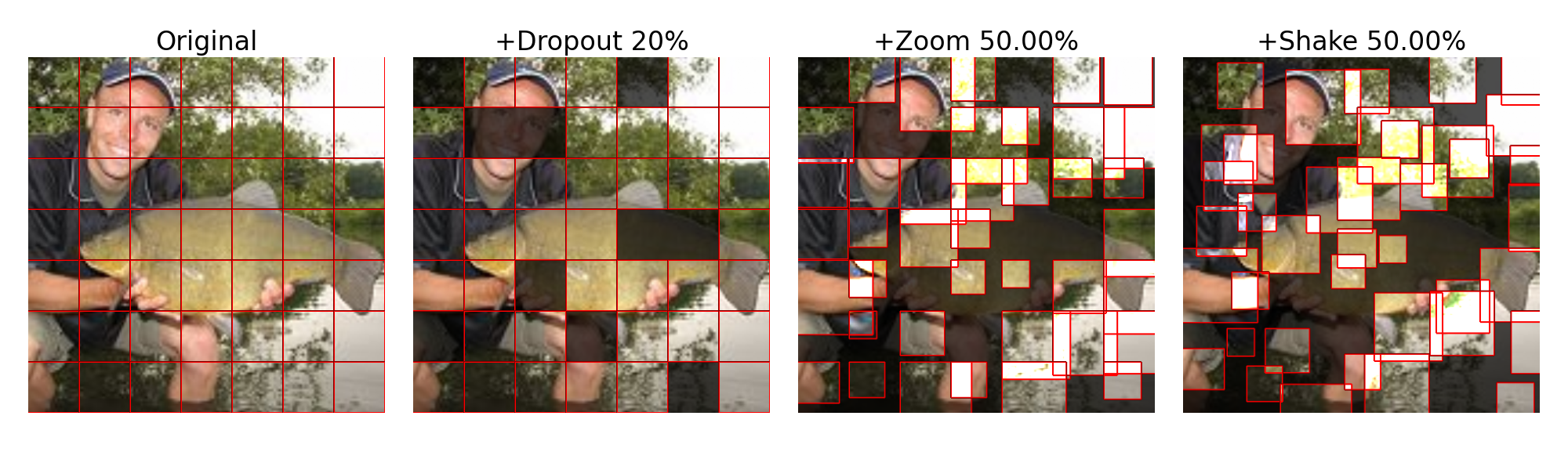}} \\
    \makebox[\textwidth][c]{\includegraphics[width=.95\textwidth]{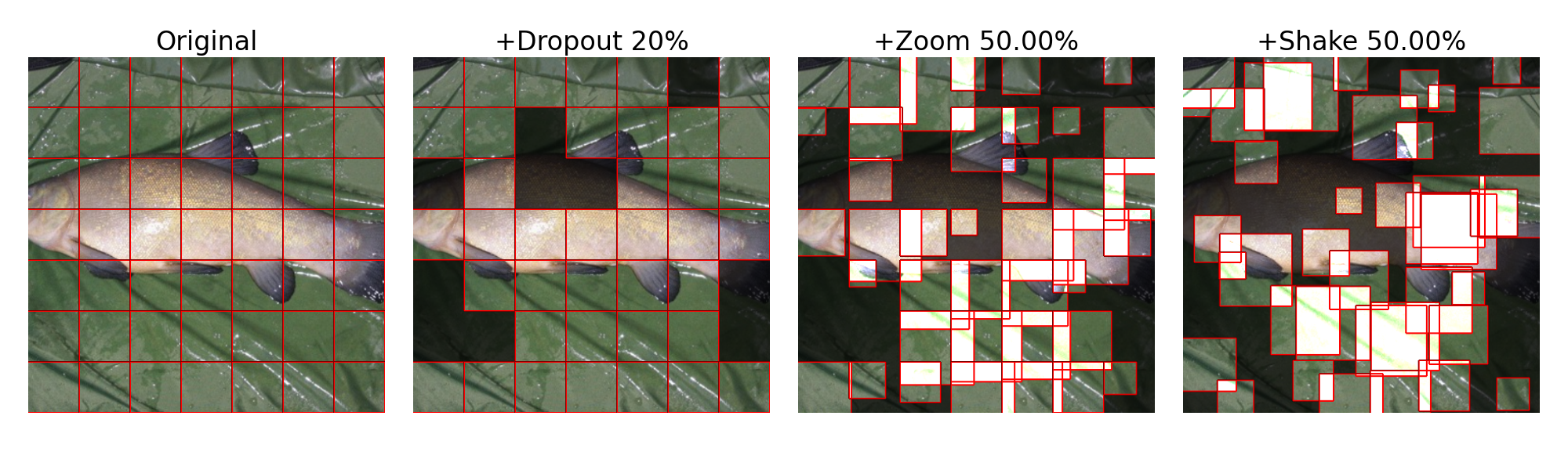}} \\ 
    \caption{\textbf{Elasticity pipeline visualization:} In this figure, we present visualizations of successive input data perturbations. For clarity, patch borders are marked in red, and patch overlap is highlighted.}
    \label{fig:elasticity-vis}
\end{figure*}

Elastic sampling allows for patch overlap, which provides redundant information to the model. In this experiment, we explored vision transformer capability to deal with redundant information. The model was provided with a standard full grid of $16\times16$ patches, that covered the entire image. Then, a number of redundant, randomly sampled patches with scales between $0.5$ and $2$ was added to the input. The results are presented in Fig.~\ref{fig:grid-redundancy}. We observe, that for standard ViT and MAE models those redundant patches essentially constitute noise, which reduces the overall performance. Our \ours{} is capable of using the extra information to slightly increase the accuracy.

\subsection{Overall performance comparison}
To assess the overall elasticity of the compared methods, in Fig.~\ref{fig:critdist} we present a critical difference diagram, aggregating results from Fig. 4 and Fig. 5 of the main paper.

\section{Pipeline visualizations}

In Fig.~\ref{fig:elasticity-vis} we present visualization of our elasticity pipeline output. For clarity, patches are separated with red borders and patch overlap is highlighted in bright colors.

\section{Theoretical analysis of input sampling strategies and their impact on positional embedding}

\subsection{Definition recall}
To strictly define the evaluation pipeline, let us consider image $I$ for which we generate set $P$ of patches $p = (x, y, s)$, where $x$ and $y$ denote the top-left corner's coordinates, and $s$ represents the relative scale (i.e. we sample a patch $r \cdot s \times r \cdot s$ and rescale it bilinearly to size $r \times r$). Initially, the coordinates $x$ and $y$ are from the regular grid and $s=1$. However, in the next step, we perturb them with three functions corresponding to the considered elasticities:
\begin{itemize}
\item $\E_{\scale(s_1, s_2)}(P)$ - introduces the scale perturbations,  sampling the $s$ parameter of every patch $p \in P$ independently and uniformly from range $[s_1, s_2]$.
\item $\E_{\miss(d)}(P)$ - adds missing data perturbations, dropping out $d$ patches from $P$ randomly with equal probability. 
\item $\E_{\pos(q)}(P)$ - applies positional perturbation, modifying $x$ and $y$ parameters of each patch $p \in P$, independently moving them by offsets sampled uniformly from range $[-r \cdot q, r \cdot q]$, where $r$ is the size of the patch.
\end{itemize}

The patches are described by their upper left $(x, y)$ and lower right corner $(x + rs, y+rs)$. Then each coordinate $\pos \in \{x, y, x+rs, y+rs\}$ is encoded by the sinusoidal positional encoding:
\begin{align*}
    \PE_{(\pos,2i)} &= \sin\left(\frac{\pos}{10000^{2i/l}}\right) \\
    \PE_{(\pos,2i+1)} &= \cos\left(\frac{\pos}{10000^{2i/l}}\right).
\end{align*}
which are concatenated into a single vector.

\subsection{Perturbations influence on positional embedding}

The application of $\E_{\miss}$ does not affect the positional encoding of a patch. However, if a patch was not dropped by the $\E_{\miss}$ perturbation, then the other two perturbations can modify its position embedding.

After application of $\E_{\pos}$, the values of $x$ and $y$ get updated to $x+ \Delta x$, $y+\Delta y$, while the lower right corner $x+rs$ and $y+rs$ get updated to values $x + rs + \Delta x$ and $y + rs + \Delta y$. The offset values $\Delta x$ and $\Delta y$ do not depend on $x$ nor $y$. Thus, for $x$, we obtain the following positional embedding
\begin{align*}
    \PE_{(x + \Delta x,2i)} &= \sin\left(\frac{x}{10000^{2i/l}} + \frac{\Delta x}{10000^{2i/l}}\right) \\
    \PE_{(x + \Delta x,2i+1)} &= \cos\left(\frac{x}{10000^{2i/l}} + \frac{\Delta x}{10000^{2i/l}}\right),
\end{align*}
which holds analogously for the other three coordinates $\{y, x+rs, y+rs\}$.

When it comes to $\E_{\scale}$, it modifies the $s$ value to $s + \Delta s$ therefore $\pos \in \{x, y\}$ remains unchanged, while both coordinates $\pos \in \{x + rs, y + rs\}$ are offset by the value $\Delta = r\Delta s$. The formula is analogous to the $\E_{\pos}$ case.